\documentclass[preprint]{elsarticle}

\usepackage{amsmath, amssymb}
\usepackage{color}
\usepackage{lineno}
\usepackage{makecell}
\usepackage{hyperref}       
\usepackage{url}            
\usepackage{amsfonts}       
\usepackage{nicefrac}       
\usepackage{microtype}      
\usepackage{doi}
\usepackage{dcolumn}
\usepackage{bm}
\usepackage{caption}
\usepackage{subcaption}
\usepackage{float}
\usepackage{array}
\newcolumntype{P}[1]{>{\centering\arraybackslash}p{#1}}

\begin{document}

\begin{frontmatter}

\title{Neural-Integrated Meshfree (NIM) Method: A differentiable programming-based hybrid solver for computational mechanics}

\author[UMN]{Honghui Du}
\author[UMN]{QiZhi He\corref{mycorrespondingauthor}}
\cortext[mycorrespondingauthor]{Corresponding author}
\ead{qzhe@umn.edu}
\address[UMN]{Department of Civil, Environmental, and Geo- Engineering, University of Minnesota, 500 Pillsbury Drive S.E., Minneapolis, MN 55455}

\begin{abstract}
While deep learning and data-driven modeling approaches based on deep neural networks (DNNs) have recently attracted increasing attention for solving partial differential equations, their practical application to real-world scientific and engineering problems remains limited due to the relatively low accuracy and high computational cost.
In this study, we present the neural-integrated meshfree (NIM) method, a differentiable programming-based hybrid meshfree approach within the field of computational mechanics.
NIM seamlessly integrates traditional physics-based meshfree discretization techniques with deep learning architectures.
It employs a hybrid approximation scheme, NeuroPU, to effectively represent the solution by combining continuous DNN representations with partition of unity (PU) basis functions associated with the underlying spatial discretization.
This neural-numerical hybridization not only enhances the solution representation through functional space decomposition but also reduces both the size of DNN model and the need for spatial gradient computations based on automatic differentiation, leading to a significant improvement in training efficiency. 
Under the NIM framework, we propose two truly meshfree solvers: the strong form-based NIM (S-NIM) and the local variational form-based NIM (V-NIM).
In the S-NIM solver, the strong-form governing equation is directly considered in the loss function, while the V-NIM solver employs a local Petrov-Galerkin approach that allows the construction of variational residuals based on arbitrary overlapping subdomains. This ensures both the satisfaction of underlying physics and the preservation of meshfree property.
We perform extensive numerical experiments on both stationary and transient benchmark problems to assess the effectiveness of the proposed NIM methods in terms of accuracy, scalability, generalizability, and convergence properties.
Moreover, comparative analysis with other physics-informed machine learning methods demonstrates that NIM, especially V-NIM, significantly enhances both accuracy and efficiency in end-to-end predictive capabilities.

\end{abstract}


\end{frontmatter}

\section{Introduction}
\subsection{Numerical methods for PDEs}
Solving partial differential equations (PDEs) is crucial to various real-world applications, particularly in material science, aerospace, civil, and mechanical engineering. 
As it is intractable to obtain analytical solutions for most complex PDEs, many canonical computational methods have been established in the past 80 years to numerically approximate the PDEs, including finite element Method (FEM)~\cite{clough1960finite,liu2022eighty}, finite volume method (FVM)~\cite{cardiff2021thirty}, meshless methods~\cite{belytschko1994element,liu1995reproducing,atluri1998new,chen2017meshfree}, and isogeometric analysis (IGA)~\cite{hughes2005isogeometric}, just to name a few.
The core idea of these approaches is to approximate both differential operators and the solution of the given PDE problem upon mesh-based or point-based spatial discretization, resulting in a set of finite-dimensional algebraic equations 
, which can be efficiently solved by modern sparse solvers. 
Many of these approaches are based on the weak form, rendering the desirable convergence property of the approximate solution.
While the above-mentioned numerical methods have demonstrated theoretical foundation and have been widely adopted in many scientific and engineering problems, 
numerous challenges persist in computational science across the disciplines such as those problems that are high-dimensional and computationally expensive, and that involve unknown (or partially unknown) physics. 

\subsection{Data-driven methods for physical simulation}
In recent decades, owing to the advancements in deep learning (DL) algorithms and increased computing power, the integration of data-driven techniques into the realm of computational science and engineering has revolutionized the way we approach complex physical problems~\cite{schmidt2009distilling,montans2019data,brunton2019data}. 
Depending on the roles that physics-based laws and data-driven techniques play in the application, this new hybrid data-physics paradigm can be broadly categorized into three classes of methods used to simulate the PDE-governing physical processes: data-fit, data-free, and blending schemes. 
The data-fit scheme can be regarded as the construction of a surrogate or reduced-order model~\cite{brunton2019data}, aiming to directly learn the underlying physics from available data, for instance, the lately emerging deep operator learning techniques approach~\cite{sanchez2020learning,lu2021learning,li2020fourier}. 
On the contrary, in the data-free methods, the existing physical laws are fully encoded/informed into the data-driven models to generate solutions by rigorously constraining the PDEs through minimizing the loss formulation, which can be thought of as an unsupervised strategy that does not require labeled data. This approach is also often referred to as physics-informed machine learning (PIML)~\cite{karniadakis2021physics,cuomo2022scientific}.
The third class refers to the approaches in which data-driven models and physical models are employed individually to approximate different components of the PDE-governing systems while they can be integrated seamlessly to perform predictive simulation. 
In the area of computational mechanics, the exemplary approaches such as model-free data-driven computing~\cite{kirchdoerfer2016data,eggersmann2019model,he2020physics_lcdd,he2021deep} and the coupled data-driven/numerical solvers~\cite{yin2022interfacing,he2023hybrid} have received extensive attention. 

\subsubsection{Physics-informed machine learning (PIML)}
Of particular interest to this study is the second class of methods, PIML, because it preserves the well-accepted physical laws whereas the prerequisite of big data is significantly relaxed. 
In this respect, one of the most prominent studies is physics-informed neural networks (PINNs)~\cite{raissi2019physics,karniadakis2021physics}, where deep neural networks (DNNs) are used to approximate the solution of PDEs with respect to space and time variables, and the strong form of the governing equations are penalized in the loss function together with the initial/boundary conditions.
While the concept of solving differential equations through neural network approximations could be traced back to at least the 1990’s~\cite{lee1990neural,meade1994numerical,lagaris1998artificial}, 
the resurgence of this PINN approach is primarily attributed to the recent advancements in deep learning infrastructure, e.g., TensorFlow, PyTorch, and JAX,
which enables efficient automatic differentiation (AD) operations~\cite{baydin2018automatic} and  optimization~\cite{goodfellow2016deep}.

PINNs provide a unified framework for solving forward, inverse, and data assimilation problems associated with PDEs~\cite{raissi2019physics, karniadakis2021physics,he2021physics,cuomo2022scientific}. 
Given its flexibility, the PINN method has been successfully applied to different engineering applications including fluid mechanics \cite{raissi2020hidden,cai2021physics}, solid mechanics \cite{haghighat2021physics,rao2021physics,yin2022interfacing}, subsurface transport~\cite{tartakovsky2020physics,he2020physics,he2021physics,du2023modeling}, among others.
Besides, various types of neural networks, such as fully connected neural networks (FCNN) \cite{raissi2019physics,berg2018unified,he2020physics}, convolutional neural networks (CNN) \cite{fang2021high,gao2021phygeonet}, 
and recurrent neural networks (RNN) \cite{zhang2020physics,taneja2023multi} have been explored to approximate the continuous solution within the PINN framework.

Although the PINN-based methods may seem to be straightforward, they usually incur difficulties in training the DNN approximation to satisfy all equation residuals, resulting in slow convergence or reduced accuracy. 
These issues have been attributed to unbalanced back-propagated gradients during model training and the notorious spectral bias of DNNs, as discussed in the studies~\cite{wang2021understanding,tancik2020fourier}.
Consequently, numerous treatments have been developed to mitigate the training issues, including adaptive weight schemes~\cite{wang2021understanding,mcclenny2020self}, domain decomposition~\cite{shukla2021parallel,kharazmi2021hp,du2023modeling}, and variable/feature transformations~\cite{wang2021eigenvector,du2023modeling}.

Nevertheless, the costly training can also be attributed to the involvement of high-order derivatives in the strong-form PDEs~\cite{krishnapriyan2021characterizing}.
Based on the weak (variational) formulation of the differential model, 
the variational PINNs (VPINNs) have been developed for solving PDEs~\cite{kharazmi2019variational,khodayi2020varnet,berrone2022variational}. Particularly, linear or high-order piecewise basis functions~\cite{khodayi2020varnet,berrone2022variational} can be implemented as the test functions in the variational framework. 
In addition, constructing the loss function based on the corresponding energy functional, deep Ritz method~\cite{yu2018deep} and deep energy method (DEM)~\cite{samaniego2020energy} were proposed for computational mechanics. 
These variational form-based PINNs can be regarded as the Petrov-Galerkin method, since the trial space approximated by DNNs is usually different from the space spanned by the test functions.
Recently, Kharazmi et al. extended the VPINN method to \textit{hp}-VPINN\cite{kharazmi2021hp} by introducing domain decomposition, allowing a localized network parameter optimization and improved training accuracy. 

\subsection{Differentiable numerical solvers}
In addition to the continuous representation of differential operators in the above-mentioned PIML approaches, leveraging physics in their discretized forms derived from classical numerical techniques has been attracting increasing attention~\cite{innes2019differentiable,kochkov2021machine,xue2023jax}. 
The recent development includes finite difference~\cite{fang2021high,chiu2022can,bezgin2023jax}, FVM~\cite{ranade2021discretizationnet}, \textit{hp} approximation~\cite{lee2021partition}, FEM~\cite{xue2023jax,dong2023deepfem}, IGA~\cite{gasick2023isogeometric}, peridynamics~\cite{haghighat2021nonlocal}, and HiDeNN~\cite{saha2021hierarchical} that can reproduce different interpolation functions based on DNN architectures. In particular, there is a growing interest to develop efficient differentiable finite discretization solvers under JAX~\cite{bradbury2018jax} such as JAX-CFD~\cite{kochkov2021machine}, JAX-Fluids~\cite{bezgin2023jax}, JAX-FEM~\cite{xue2023jax}, and JAX-DISP~\cite{mistani2023jax}. 
It is noted that all these discretization-based solvers are based on differentiable programming~\cite{baydin2018automatic,innes2019differentiable}, which embeds the numerical linear algebra and gradient operations in the neural network architectures, enabling the application of end-to-end differentiable gradient-based optimization methods. 

Despite the simplicity of implementing these differentiable numerical solvers and their potential for significant acceleration on specialized hardware, this field is still in its nascent stages and requires further development. Specifically, two limitations should be highlighted.
First, as pointed out in~\cite{johnson2023software,mistani2023jax}, the fundamental assumption that AD capabilities within current machine learning frameworks can compute “exact” derivatives through complex architectural neural network models may not always be valid.
This can result in inaccuracies in the spatial gradients of partial differential equations (PDEs) or discretized differential operators during the training of numerical models, leading to suboptimal convergence and potentially unstable approximations~\cite{johnson2023software, krishnapriyan2021characterizing}.
Second, many of the aforementioned differentiable numerical methods rely on structured or conforming grids to approximate gradients, which compromises the truly \textit{meshfree} property, 
a distinctive feature often associated with PIML methods.
This contradicts the applicability advantages offered by modern neural networks.
\subsection{Differentiable meshfree solver}
To overcome these limitations, we develop a differentiable-programming hybrid method, which incorporates the conventional physics-based meshfree discretization method in a deep learning architecture to achieve highly accurate, efficient, and end-to-end training and predictive simulation.
Furthermore, motivated by the advantages of nodal shape functions employed in traditional numerical methods~\cite{babuvska1997partition,belytschko1994element,liu1995reproducing,atluri1998new,hughes2005isogeometric}, while simultaneously harnessing the universal approximation capabilities of DNNs, we introduce an auxiliary hybrid approach for solution approximation.
This hybrid scheme, termed the \textit{Neuro-partition of unity} (NeuroPU) approximation, is to interpolate the solution by seamlessly integrating a set of partition of unity (PU) basis functions~\cite{babuvska1997partition} defined on the underlying meshfree discretization with DNN represented nodal coefficient functions.
This new framework is coined the \textit{neural-integrated meshfree} (NIM) method.

In this work, without loss of generality, we chose to employ the reproducing kernel (RK) meshfree shape functions~\cite{liu1995reproducing,chen1996reproducing} in the NeuroPU approximation since the RK shape functions, constructed based on spatially distributed nodes over physical domain, offer the flexibility to design arbitrary order of accuracy, smoothness, and compactness.
The motivation of introducing NeuroPU for solution approximation is to regulate the solution representation by well-established PU basis functions and mitigate the need for intricate computations of high-order gradients that usually rely on AD, ultimately improving training efficiency.
On the other hand, NeuroPU leverages embedded neural networks to represent the functional space related to problem-related parameters, enabling effective surrogate modeling of parameterized or time-dependent problems. 

In this study, we will present two NIM solvers for computational mechanics modeling: strong form-based NIM (S-NIM) and local variational form-based NIM (V-NIM). 
Notably, both of these solvers are genuinely meshfree, eliminating the need for high-cost conforming mesh generation.
Like the PINN methods~\cite{raissi2019physics,karniadakis2021physics,he2021physics,cuomo2022scientific}, the former one considers the strong-form governing equations in the associated loss function. We will demonstrate the superior performance of S-NIM over standard PINNs due to the incorporation of NeuroPU approximation, which substantially reduces the DNN solution space and improves the training efficiency and accuracy.
In order to further improve the accuracy and stability, we propose the V-NIM solver, which is inspired by the meshless local Petrov–Galerkin method (MLPG)~\cite{atluri1998new,atluri2000new} that drives the consistent weak formulation over local subdomains. 
Distinct from other variational PINN methods~\cite{khodayi2020varnet,kharazmi2021hp,berrone2022variational}, the proposed V-NIM allows for using arbitrary overlapping subdomains to formulate the loss function, and thus, upholding the meshfree property.
We demonstrate the outstanding performance of the proposed NIM methods, especially V-NIM, through extensive experiments on the benchmark examples (e.g., Poisson equation, linear elasticity, time-dependent problem, and a parameterized PDE), where the accuracy, convergence property, generalizability, and efficiency are compared against other baseline methods. 
To the best of the authors’ knowledge, this study represents the first attempt to 
develop differentiable programming-based meshfree solvers integrated with hybrid neuro-numerical approximation for computational mechanics modeling.

The remainder of the paper is organized as follows: Section \ref{sec:prelim} provides a background review of numerical discretization and DNNs for function approximation. In Section \ref{sec:nma}, we delve into the construction of hybrid NeuroPU approximation. We present the methodology development of the NIM framework based on the NeuroPU approxiamtion and meshfree discretization in Section \ref{sec:NIM},
followed by the detailed solution procedures of the proposed S-NIM and V-NIM methods provided in Section \ref{sec:procedure}.
Numerical tests on static and time-dependent benchmark problems are presented in Section \ref{sec:result} and Section \ref{sec:result_ade}, respectively. 
Section \ref{sec:conclusion} concludes the paper by summarizing the main findings and contributions.

\section{Preliminaries}\label{sec:prelim}

In this section, we first present a background review of the two fundamental methodologies of function approximation based on numerical discretization and deep neural networks (DNNs), as well as their applications to solving PDEs in computational mechanics.

For demonstration, a classical linear elastostatics problem is taken as the model problem.
Let us consider an elastic solid defined in a bounded domain $\Omega \subset \mathbb{R}^d$, where $d$ denotes the spatial dimension, and its boundary $\partial \Omega \subset \mathbb{R}^{d-1}$ is split as the essential boundary condition (EBC) on $\Gamma_g $ and the natural boundary condition (NBC) on $\Gamma_t$, i.e., $\partial \Omega = \Gamma_g \cup \Gamma_t$ with $\Gamma_g \cap \Gamma_t = \emptyset $. The governing PDE describing the static equilibrium is written as:
\begin{equation} \label{eq:elastic}
\begin{cases}
\nabla \cdot \boldsymbol{\sigma}(\bm u)+\boldsymbol{f}= \bm 0, \quad & \text{in} \quad \Omega \\ 
\boldsymbol{n} \cdot \boldsymbol{\sigma}(\bm u)=\overline{\boldsymbol{t}}, \quad & \text{on} \quad \Gamma_t \\
\bm u ={\overline{\bm u}}, \quad & \text{on} \quad \Gamma_g
\end{cases}
\end{equation}
where $\bm u$ is the displacement vector, $\bm \sigma$ is the Cauchy stress tensor, $\bm f$ is the body force, $\overline{\bm u}$ and $ \overline{\bm t}$ are the displacement and traction values prescribed on
$\Gamma_g $ and $\Gamma_t$, respectively, and $\bm n$ is the surface normal on $\Gamma_t$. 

For solid mechanics problems, the constitutive law that relates stress and strain, e.g., $\bm \sigma = \bm \sigma (\bm \varepsilon(\bm u))$, is required to solve the boundary value problem (BVP) in \eqref{eq:elastic}, where the linear strain tensor is given by
\begin{equation}\label{eq:elst_strain}
\bm{\varepsilon} := \nabla_{sym} \bm{u} = \frac{1}{2}(\nabla \boldsymbol{u}+ \nabla \bm u^T)
\end{equation}
For linear elastic materials, the strain-stress relation becomes
\begin{equation}
\bm \sigma=\boldsymbol{C}: \bm \varepsilon
\label{eq:stress}
\end{equation}
where $\bm C$ is the elasticity tensor.

For many engineering applications, such as inverse design, uncertainty quantification, and surrogate modeling, varying model parameters are considered in the solid mechanics analysis \eqref{eq:elastic}. Here, $\bm{\mu} \in \mathbb{R}^{c}$ is used to denote the parameter vector associated with the problem-specific variables,
such as materials properties, loading, and boundary conditions, etc. In this scenario, the displacement solution is then generalized as a parameter-dependent field $\bm{u} (\bm{x},\bm{\mu}): \Omega \times \mathbb{R}^{c}  \mapsto \mathbb{R}^{d}$.

\subsection{Numerical approximation via shape functions}\label{sec:shape_function}

In spatial discretization-based numerical methods for solid mechanics problems (e.g., FEM~\cite{hughes2012finite}, meshfree methods~\cite{belytschko1994element,liu1995reproducing}, 
and IGA~\cite{hughes2005isogeometric}), the cornerstone is to apply nodal shape functions to approximate the solution field in a finite-dimensional manner. 

Let the computing domain $\Omega$ be discretized by a set of nodal points $\{\bm x_I\}_{I=1}^{N_h}$, the approximation of solution $\bm{u}^h$ is defined as the linear combination of nodal shape functions:
\begin{equation}\label{eq:pu_1}
\bm u (\bm{x}) \approx \bm{u}^h (\bm{x}) = \sum_{I=1}^{N_h} \Psi_I(\bm x) \bm{d}_I
\end{equation}
where $N_h$ represents the number of nodes, $\bm{d}_I \in \mathbb{R}^d$ is the nodal coefficient at location $\bm{x}_I$, and $\Psi_I$ is the shape function associated with the $I$th node. 

While the shape functions $\Psi_I$ can be defined in many different forms, 
they are required to satisfy specific conditions (completeness and continuity) to ensure the convergence property of approximate solutions. 
One essential condition is the so-called partition of unity (PU)~\cite{babuvska1997partition}, which requires that the ensemble of the compact support of shape functions generates a covering for domain $\Omega$, i.e., $\Omega \subset \bigcup_{I=1}^{N_h} \operatorname{supp}\{\Psi_I\}$, and $\sum_{I=1}^{N_h} \Psi_I = 1$ with $0 \le \Psi_I (\bm x) \le 1$.

For $\bm{x} \in \Omega$, let $\mathcal{S}_x=\left\{I \mid \boldsymbol{x} \in \operatorname{supp}\left(\Psi_I\right)\right\}$ be an index set of the nodal shape functions whose influence do not vanish at location $\bm x$, the PU approximation \eqref{eq:pu_1} can be rewritten as
\begin{equation}\label{eq:pu_2}
\bm{u}^h (\bm{x}) = \sum_{I \in \mathcal{S}_x} \Psi_I(\bm x) \bm{d}_I
\end{equation}
In the following, we introduce a special type of meshfree shape functions, i.e., reproducing kernel (RK) approximation, which allows an arbitrary order of accuracy while maintaining higher-order smoothness~\cite{liu1995reproducing,chen1996reproducing,chen2017meshfree}.
The RK shape functions $\{\Psi_I\}_{I=1}^{N_h}$ used for approximating displacement take the following form:
\begin{equation}\label{eq:RK_shape}
\Psi_I(\boldsymbol x)=\boldsymbol p^{[p] T}\left(\boldsymbol x_I-\boldsymbol x\right) \bm{b}(\boldsymbol x) \phi_a\left(\boldsymbol x_I-\boldsymbol x\right)
\end{equation}
where the kernel function $\phi_a$ controls the smoothness of the RK approximation function and defines the compact support with a size $a$.
A widely used kernel function is the cubic B-splines that preserves $C^2$ continuity
\begin{equation}\label{eq:kernel_B}
\phi_a(z)= \begin{cases}\frac{2}{3}-4 z^2+4 z^3 & 0 \le z \leq \frac{1}{2} \\ 
\frac{4}{3}-4 z+4 z^2-\frac{4}{3} z^3 & \frac{1}{2}< z\leq 1 \\
0 & z > 1 
\end{cases}
\end{equation}
with $z=\lVert \boldsymbol x_I-\boldsymbol x \rVert / a$.
In \eqref{eq:RK_shape}, $\boldsymbol{p}^{[p]}(\boldsymbol{x})$ is a vector of monomial basis functions up to the $p$th order
\begin{equation}\label{eq:basis}
\boldsymbol p^{[p]}(\boldsymbol x)=\left\{1, x_1, x_2, x_3, \cdots, x_1^i x_2^j x_3^k, \ldots, x_3^p\right\}^T, 0 \leq i+j+k \leq p
\end{equation}
and the parameter vector $\boldsymbol{b}(\boldsymbol{x})$ is determined by enforcing the following $p$th order reproducing conditions:
\begin{equation}
\sum_{I \in \mathcal{S}_x} \Psi_I(\boldsymbol{x}) \boldsymbol{p}^{[p]}\left(\boldsymbol x_I\right)=\boldsymbol{p}^{[p]}(\boldsymbol{x})
\label{eq:repro}
\end{equation}
Substituting Eq. \eqref{eq:repro} into Eq. \eqref{eq:RK_shape} yields
\begin{equation}
\bm{b}(\boldsymbol x)=\boldsymbol A^{-1}(\boldsymbol{x}) \boldsymbol{p}^{[p]}(\boldsymbol{0})
\label{eq:C}
\end{equation}
where $\bm{A}(\bm{x})$ is the moment matrix
\begin{equation}
\bm A(\boldsymbol x)=\sum_{I \in \mathcal{S}_x} \boldsymbol p^{[p]}\left(\boldsymbol x_I-\boldsymbol x\right) \boldsymbol p^{[p] T}\left(\boldsymbol x_I-\boldsymbol x\right) \phi_a \left(\boldsymbol x_I-\boldsymbol x\right)
\label{eq:A}
\end{equation}
Invoking \eqref{eq:C} into \eqref{eq:RK_shape}, we have the following expression for the RK shape functions:
\begin{equation}
\Psi_I(\boldsymbol x)=\boldsymbol p^{[p] T}(\boldsymbol x) \boldsymbol A^{-1}(\boldsymbol x) \boldsymbol p^{[p]}\left(\boldsymbol x_I-\boldsymbol x\right) \phi_a\left(\boldsymbol x_I-\boldsymbol x\right)
\end{equation}
Subsequently, the nodal coefficients in Eq. \eqref{eq:pu_2} can be determined by solving the Galerkin weak formulation~\cite{hughes2012finite} corresponding to the elasticity problem \eqref{eq:elastic}.

Specifically, the RK shape function with quadratic basis function ($p=2$) and the cubic B-spline function \eqref{eq:kernel_B} is, by default, employed in the following numerical study, which will be further discussed in Section \ref{sec:nma} and \ref{sec:result}. Additionally, we denote a normalized support size with the characteristic nodal distance $h$ as $\bar a=a/h$.

\subsection{Neural network-based approximation for solving PDEs}\label{sec:DNN}

Thanks to the universal approximation~\cite{hornik1991approximation, blum1991approximation},
DNNs are capable of approximating arbitrary continuous functions, which makes it become an emerging candidate of approximate solutions in solving PDEs. 
By taking the spatial coordinates as inputs, denoted as $\mathbf z_0 = \bm x$,  we consider a feed-forward fully-connected neural network (FFCN) which consists of the input layer, $n$ hidden layers, and the output layer, defined as
\begin{equation}
    \boldsymbol{u}(\mathbf z_0 ) \approx \hat{{\boldsymbol{u}}}(\mathbf z_0; \bm{\theta})=\mathbf {z}_{n+1}\left(\mathbf {z}_n\left(\cdots \mathbf {z}_2\left(\mathbf {z}_1(\mathbf {z_0})\right)\right)\right)
\end{equation}
in which the symbol $\string ^$ is used to indicate a variable that is parameterized by DNNs, and $\bm{\theta}$ represents the collection of trainable parameters associated with the DNN model. The connection between $(l-1)$th layers and $l$th layer is defined as
\begin{equation}
\begin{gathered}
\mathbf{z}_l=\mathbf{z}_l (\mathbf{z}_{l-1})=\sigma\left(\mathbf{b}_l+\mathbf{W}_l \mathbf{z}_{l-1}\right), \quad \text{for } 1 \leq l \leq n \\
\mathbf{z}_{n+1}=\mathbf{b}_{n+1}+\mathbf{W}_{n+1} \mathbf{z}_n, \quad \text{for } l=n
\end{gathered}
\end{equation}
where the activation function $\sigma(\cdot)$ is selected as hyperbolic tangent function, $\mathbf{z}_{n+1}$ denotes the output vector, and $\mathbf W_i$ and $\mathbf b_i$ correspond to the weights and bias associated with the $i$th layer, respectively.

\subsubsection{PINNs}
When using the PINN method~\cite{raissi2019physics,karniadakis2021physics} for solving the elasticity problem in Eqs. \eqref{eq:elastic}-\eqref{eq:stress}, a DNN model is used to approximate the displacement solution, denoted as $\hat{\bm u}(\bm x)$, with the spatial coordinates $\bm x$ as the network inputs.
The differential operators, e.g., $\nabla \hat{\bm u}(\bm x; \bm \theta)$, are computed by performing Automatic Differentiation (AD) \cite{baydin2018automatic} with respect to the inputs. 
As a result, the mean square errors (MSEs) of the governing partial differential equations (PDEs) are subsequently incorporated into the loss function by feeding both the approximate solution and its derivatives on the given set of collocation points. 
With slightly abusing the notion, the corresponding loss of the elasticity problem in \eqref{eq:elastic}-\eqref{eq:stress} is given as
\begin{equation}
\begin{aligned}
\mathcal{L}(\bm \theta) = | \hat{\bm u} - \overline{\boldsymbol{u}}|_{\Gamma_{g}} 
+ | \boldsymbol{n} \cdot (\boldsymbol{C}: \nabla_{sym} \hat{\bm u}) - \overline{\boldsymbol{t}}|_{\Gamma_{t}}
+ |\nabla \cdot (\boldsymbol{C}: \nabla_{sym} \hat{\bm u}) +\boldsymbol{f}|_{\Omega}
\end{aligned}
\label{eq:loss_PINN}
\end{equation}
The parameters $\bm \theta$ of neural networks are optimized through the minimization of the loss function. For a more formal formulation regarding the standard PINN~\cite{haghighat2021physics} or the mixed-form PINN~\cite{rao2021physics} for linear elasticity, we refer readers to the following studies~\cite{raissi2019physics,haghighat2021physics,rao2021physics,he2021physics}.

Aided by DNNs, the traditional PINN methods offer a straightforward way to approximate the solution fields by incorporating the given physical laws.
Nevertheless, the complex solution fields often necessitate a relatively large size of neural networks to ensure sufficient approximation capacity, inevitably leading to over-parameterized search space that poses challenges in training the non-convex PDEs-informed loss functions~\cite{he2020physics, cuomo2022scientific, chiu2022can}.
Consequently, a huge amount of collocation points as well expensive optimizer iterations are usually required to train the PINN model resulting from complex high-dimensional PDE problems~\cite{cuomo2022scientific,krishnapriyan2021characterizing,du2023modeling}.
Moreover, performing AD operators with regard to large data points over spatiotemporal domain usually takes a considerable portion of computational time, especially when dealing with high-order PDEs.

\section{Hybrid approach: Neuro-partition of unity (NeuroPU) approximation}\label{sec:nma}
\begin{figure}[htb]
	\centering
    \includegraphics[angle=0,width=1.0\textwidth]{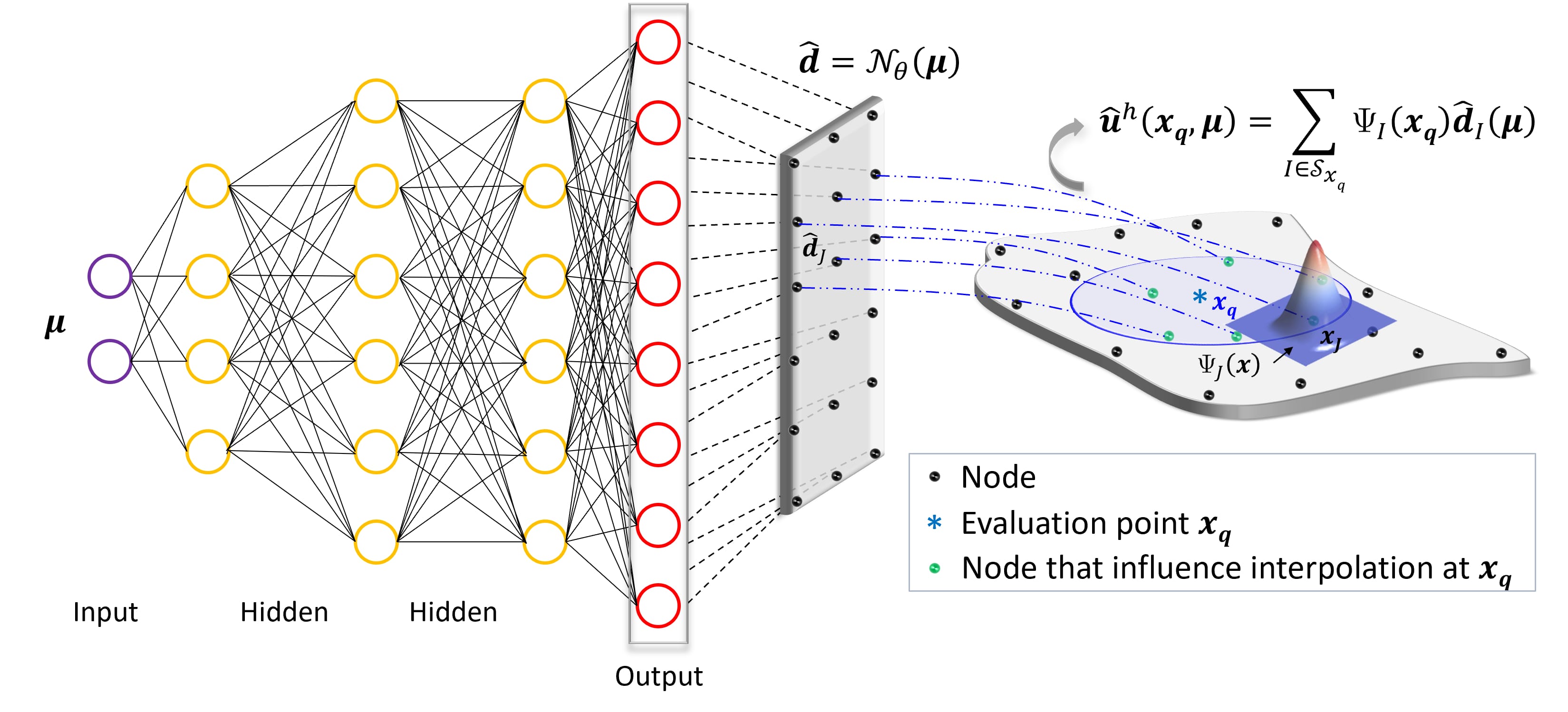}
	\caption{
 Schematic of neuro-partion of unity (NeuroPU) approximation \eqref{eq:nmm}, 
    where $\Psi_I$ is the nodal shape function centered at $\bm x_I$, $I=1,2,...N_h$, related to the numerical discretization over the physical domain,
    and the coefficient functions $\hat{\bm{d}}\in \mathbb{R}^{d \times N_h}$ are approximated by 
    using neural networks with the system parameters $\bm{\mu} \in \mathbb{R}^{c}$ as inputs. The sizes of input layer and output layer of neural network are decided by the dimensions of $\bm \mu$ and $\hat{\bm{d}}$, respectively. }
    \label{fig:structure}
\end{figure}

In order to enhance the training process and accuracy within the field of PIML, we propose a hybrid approach called neural-partition of unity (NeuroPU) approximation, which integrates numerical discretization generated by PU shape functions (Section \ref{sec:shape_function}) over the physical domain and the neural network-based approximation (Section \ref{sec:DNN}) for nodal coefficient functions. This hybrid approximation method is the building block of the proposed NIM framework, which will be further elaborated in Section \ref{sec:NIM}.

Given a set of predefined nodal shape functions, the NeuroPU approximation of the parametric solution $\bm{u}(\bm x, \bm \mu)$ is expressed as
\begin{equation}
\hat{\bm{u}}^h(\boldsymbol{x},\boldsymbol \mu) =  \sum_{I \in \mathcal{S}_x} \Psi_I(\bm{x}) \hat{\bm d}_I(\bm \mu)
\label{eq:nmm}
\end{equation}
where $\Psi_I$ is the PU shape function and specifically the RK shape function defined in \eqref{eq:RK_shape} is adopted, and $\mathcal{S}_x$ is the set of nodes that contribute to the interpolation at $\bm x$ (see Section \ref{sec:shape_function}). 
In Eq. \eqref{eq:nmm}, the nodal coefficient function $\hat{\bm{d}}_I (\bm\mu) \in \mathbb{R}^d$ is the corresponding $I$th output of a neural network $\mathcal{N}_{\theta}$ which is parametrized by $\bm \theta$ as explained in Section \ref{sec:DNN}. Let the collection of nodal coefficients be denoted as $\hat{\bm{d}}:= \{\hat{\bm{d}}_I\}_{I=1}^{N_h} \in \mathbb{R}^{d \times N_h} $, which defines a mapping from inputs $\bm{\mu}$ to the discrete nodal variable $\hat{\bm{d}}$ through a multi-layer neural network, i.e.,
\begin{equation}\label{eq:nma_dnn}
\hat{\bm{d}}: \bm{\mu} \in \mathbb{R}^{c}  \mapsto \mathcal{N}_{\theta}(\bm{\mu})
\end{equation}
Notably, the utilization of the DNN model allows the nodal coefficients to be expressed as a function of problem-defined system parameters  $\boldsymbol{\mu}$, e.g., temporal coordinates and material coefficients. 
Thus, the NeuroPU approximation can be readily employed for a surrogate model for parameterized systems, as demonstrated in Section \ref{sec:surrogate}.

Due to the compact nature of the support domain, only a few entries of $\Psi_I$ and $\hat{\bm{d}}_I$ interacted within the support domain $\mathcal{S}_x$ will be active in the dot product in Eq. \eqref{eq:nmm}. This results in a sparsity structure that can streamline matrix calculations and save storage.
Besides, the hybrid NeuroPU approximation offers an efficient computation of all the space-dependent derivative terms.
To wit, since spatial coordinates are only involved via the shape functions in the NeuroPU construction, the spatial gradient of $\hat{\bm u}^h$ yields
\begin{equation}\label{eq:grad_nma}
\nabla {\hat{\bm u}}^h(\bm{x},\bm \mu) =  \sum_{I \in \mathcal{S}_x} \nabla \Psi_I(\bm{x}) \hat{\bm d}_{I}(\bm \mu)
\end{equation}
In this setting, the derivatives of shape functions, $\nabla \Psi_I$, can be pre-computed and stored in advance for the subsequent derivation of  $\nabla {\hat{\bm u}}^h$ when establishing the loss function. This is essentially distinct from the PINN method where the differential and gradient operators in governing equations are computed through automatic differentiation during training.
Furthermore, in contrast to the studies~\cite{lee2021partition,saha2021hierarchical}, where shape functions are implicitly encoded using architectural or hierarchical neural networks, the NeuroPU approximation explicitly expresses these shape functions, which preserves the necessary simplicity and compatibility for scalable implementation, akin to classical numerical methods.

\emph{Remark 3.1.} It is noted that while the reproducing kernel (RK) approximation is adopted as PU shape functions in Eq. \eqref{eq:nmm}, other types of PU shape functions
, e.g., Lagrange polynomial basis~\cite{clough1960finite,hughes2012finite}, spectral series \cite{patera1984spectral}, and NURBS~\cite{hughes2005isogeometric}, are also applicable to the proposed framework. 
However, to ensure truly meshfree properties and avoid tedious mesh generation, we adopt meshfree-type shape functions that are defined on physically spatial domain and allow overlapping compact supports. 

\emph{Remark 3.2.}
The proposed NeuroPU approach offers the flexibility to utilize various parameter inputs $\bm \mu$ and neural network architectures for nodal coefficient functions $\hat{\bm d}_I(\bm \mu)$, depending on the specific problem at hand. For example, when dealing with temporal variable as input, fully connected neural networks (FCNN) provide a continuous representation that captures temporal evolution, while convolutional neural networks (CNN) can be employed to process the image inputs, obtaining nodal coefficient functions with discrete representations.

\section{Neural-Integrated Meshfree (NIM) Framework}\label{sec:NIM}

This section aims to develop the \textit{neural-integrated meshfree (NIM)} method, a novel differentiable programming-based meshfree computational framework for solving computational mechanics problems. Particularly, we will focus on establishing a highly efficient neural-numerical platform by leveraging both the hybrid NeuroPU approximation and meshfree formulation that is well designed for the end-to-end neural network training procedure.

In the following subsections, we propose two NIM solvers that are formulated based on the strong form and the local variational form of governing equations, denoted as S-NIM and V-NIM, respectively.

\subsection{Approach I: Strong form-based neural integrated meshfree solver (S-NIM)} \label{sec:s-nim}

\begin{figure}[htb]
	\centering
	\includegraphics[angle=0,width=1.0\textwidth]{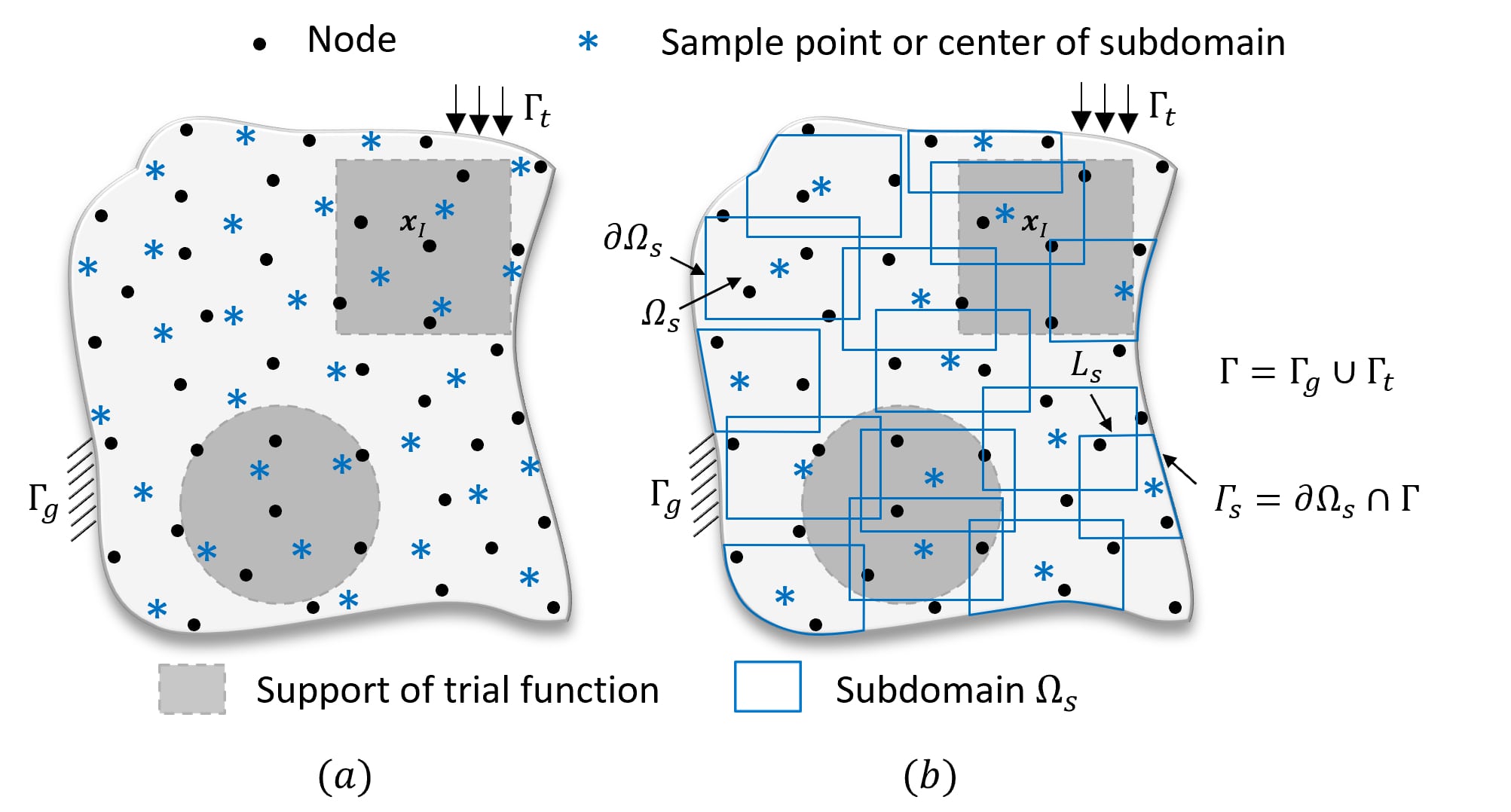}
	\caption{Schematics of the S-NIM and V-NIM methods, where the nodes are represented by black points, and the influence domains $supp(\bm x_I)$ of the trial shape functions ($\Psi_I$) are indicated by grey rectangular or square areas around nodes. (a) S-NIM: The sample points are represented by blue stars; (b) V-NIM: The local subdomains, denoted by $\Omega_s$, are represented by blue rectangles, with their centers marked as blue stars. 
    The boundary of the local subdomain, $\partial \Omega_s$, is divided into two parts, $\Gamma_{s}$ and $L_s$, where $\Gamma_{s}$ represents the portion of $\partial \Omega_s$ that lies on the global boundary $\partial \Omega$, whereas $L_s$ corresponds to the portion of the boundary located within the domain $\Omega$.}
\label{fig:NIM}
\end{figure}

With the employment of the NeuroPU approximation \eqref{eq:nmm}, we first develop the strong form-based NIM solver (S-NIM), which is designed to directly encode the governing equations and boundary conditions associated with the problem in the network structure.
Similar to the PINN method described in \eqref{eq:loss_PINN},
the loss function in S-NIM is formulated as the sum of mean squared residuals of Eqs. \eqref{eq:elastic}-\eqref{eq:stress}:
\begin{equation}
\begin{aligned}
\mathcal{L}^S(\bm \theta) = \frac{1}{N_{\mu}} \sum_{j=1}^{N_{\mu}} \left\{ \frac{1}{N_f} \sum_{i=1}^{N_f} \left\| \nabla \cdot \hat{\bm \sigma}^h\left(\bm x_i , \bm\mu_j\right) + \boldsymbol{f}\left(\bm x_i , \bm\mu_j\right) \right\|^2  \right. \\
+ \frac{\alpha_1}{N_t} \sum_{i=1}^{N_t} \left\| \hat{\boldsymbol{t}}^h\left(\bm x_i , \bm\mu_j\right) - \overline{\boldsymbol{t}}\left(\bm x_i , \bm\mu_j\right) \right\|^2  \\
+ \left. \frac{\alpha_2}{N_g} \sum_{i=1}^{N_g} \left\| \hat{\boldsymbol{u}}^h\left(\bm x_i , \bm\mu_j\right) - \overline{\boldsymbol{u}}\left(\bm x_i , \bm\mu_j\right) \right\|^2 \right\}
\end{aligned}
\label{eq:loss_snim}
\end{equation}
where the matrix forms of $\hat{\bm u}^h$, $\hat{\bm \sigma}^h$ and $\hat{\bm t}^h$ are provided in \ref{sec:app_A}. 
The superscript $h$ denotes the variables involving the NeuroPU approximation. 
$N_f$ represents the number of residual points $\mathcal{S}_f = \{\bm x_i\}_{i=1}^{N_f}$ sampled over computation domain $\Omega$ (denoted by blue stars in Figure \ref{fig:NIM}a), $N_t$ and $N_g$ are the numbers of sampling points $\mathcal{S}_t = \{\bm x_i\}_{i=1}^{N_t}$  and $\mathcal{S}_g = \{\bm x_i\}_{i=1}^{N_g}$ whereby EBC on $\Gamma_g$ and NBC on $\Gamma_t$ are imposed, respectively. 
Here, we consider a mechanics problem parameterized by the parameter vector $\bm \mu$, and $N_{\mu}$ denotes the number of sample points in the parameter set, i.e., $\mathcal{S}_\mu = \{\bm \mu_i\}_{i=1}^{N_\mu}$. 

In Eq. \ref{eq:loss_snim}, the weight coefficients $\alpha_1$ and $\alpha_2$ are used to penalize the loss terms associated with boundary conditions. It has been reported that the weights are critical to the convergence rates of loss functions and the accuracy of approximation solution \cite{he2021physics, wang2021understanding}. 
While the loss function of S-NIM resembles the standard PINN method for elasticity problems~\cite{haghighat2021physics,rao2021physics},
distinct approximation functions are used.
Our numerical studies in Section \ref{sec:result} will demonstrate that the introduction of the NeuroPU approximation in S-NIM significantly boosts both the training efficiency and accuracy, 
attributing to the reduced dimensionality in approximation space and the efficient, high-order accurate spatial gradients provided by RK shape functions.

\subsection{Approach II: Local variational form-based neural integrated meshfree framework (V-NIM)}\label{sec:v-nim}

While formulating the loss function for the proposed S-NIM method is straightforward due to the employment of strong-form governing PDEs, the involvement of higher-order derivatives in loss functions could potentially impact the efficiency and accuracy. 
Several \textit{variational} physics-informed machine learning methods~\cite{khodayi2020varnet, kharazmi2021hp, berrone2022variational}, as discussed in the introduction, have been developed to alleviate these issues,
where the corresponding loss function is constructed based on the variational (weak) form of the governing equations derived by using the weighted residual methods along with various testing functions.
The use of weak form decreases the required regularity of the approximate solution. 
Consequently, a reduction in the highest order of derivatives and an improvement in training accuracy can be achieved when compared to the strong form counterpart.

However, these variational approaches typically require a conforming discretization to construct the loss function described by an integral form on the entire computation domain,
which inevitably results in the loss of the truly meshless characteristic.
Motivated by this concern,
we propose to introduce a local weak formulation in the NIM method, namely V-NIM, which effectively preserves the discretization/mesh-free property. 
This V-NIM approach admits the local incorporation of underlying physics and avoids the significant costs associated with mesh generation.

\subsubsection{Local weak formulation for BVP}\label{sec:localweak}

Inspired by the idea of local weak form used in the meshless local Petrov-Galerkin (MLPG) approach \cite{atluri1998new,atluri2000new, han2004meshless},
we let $\mathcal{T}$ be the set of (overlapping) local subdomains such that their union covers the whole domain, i.e., $\bar{\Omega} \subset \bigcup_{s \in \mathcal{T}} \{\Omega_s\}$, with $N_{\mathcal{T}} = |\mathcal{T}|$ denoting the number of subdomains in $\mathcal{T}$, as shown in Figure \ref{fig:NIM}b.
Under this setting, we can define an arbitrary local test function $\bm{v}$ on the subdomain $\Omega_s$, i.e., $\bm{v}(\bm x): \bm{x} \in \Omega_s \subset \Omega \mapsto \mathbb{R}^d$. Therefore, the local weak form of the governing equation \eqref{eq:elastic} over $\Omega_s$ is written as
\begin{equation}
\int_{\Omega_s} \boldsymbol{v} \cdot\left(\nabla \cdot \boldsymbol{\sigma}+\boldsymbol{f}\right) d \Omega = 0
\end{equation}
Applying integration by parts and the divergence theorem to the above equation, with introducing the traction/natural boundary condition, yields the following local variational (weak) formulation
\begin{equation}
\begin{aligned}
\int_{\Omega_s} \nabla \boldsymbol{v}: \boldsymbol{\sigma} d \Omega 
- \int_{L_s} \boldsymbol{v} \cdot \bm{t} d \Gamma
- \int_{\Gamma_{s g}} \boldsymbol{v} \cdot \bm{t} d \Gamma \\
= \int_{\Omega_s} \boldsymbol{v} \cdot \boldsymbol{f} d \Omega + \int_{\Gamma_{s t}} \boldsymbol{v} \cdot \overline{\boldsymbol{t}} d \Gamma 
\end{aligned}
\label{eq:local_weak}
\end{equation}
where $\bm{t} = \bm{\sigma} \cdot \bm{n}$. In general, the boundary of the local domain $\Omega_s$ is $\partial \Omega_s = \Gamma_s \cup L_s$, in which $\Gamma_{s}$ denotes the portion of the local boundary $\partial \Omega_s$ located on the global boundary $\Gamma = \partial \Omega$, i.e., $\Gamma_s = \partial \Omega \cap \Gamma$, whereas $L_s$ represents the remaining part of the local boundary that lies inside the domain, as shown in Figure \ref{fig:NIM}b.
Specifically, we denote $\Gamma_{s g} = \Gamma_s \cap \Gamma_g$ and $\Gamma_{s t} = \Gamma_s \cap \Gamma_t$ as the parts of the local boundary $\Gamma_s$ on which the EBC and NBC are specified, respectively. For a subdomain located entirely within the global domain, there exists no $\Gamma_s$, and thus, the boundary integrals over $\Gamma_{s g}$ and $\Gamma_{s t}$ vanish and $L_s = \partial \Omega_s$.
It should be emphasized that the essential boundary conditions have not yet been imposed in Eq. \eqref{eq:local_weak}, and this will be addressed in Section \ref{sec:EBC}.

Given a selected test function $\bm{v}$ defined on $\Omega_s$, the discretization of Eq. \eqref{eq:local_weak} will only yield one linear algebraic equation. To ensure obtaining a sufficient number of linearly independent equations for displacement solution $\bm{u} \in \mathcal{R}^d$, we can apply $N_v$ $(N_v \ge d)$ independent sets of test functions $\{ \bm{v}^{(k)} \}_{k=1}^{N_v}$ to Eq. \eqref{eq:local_weak}. As such, the local variational residual associated with the $k$th test function $\bm{v}^{(k)}$ over $\Omega_s$
is defined as
\begin{equation}
\begin{aligned} \mathcal{R}_{s}^{(k)} & =\int_{L_s} \boldsymbol{v}^{(k)} \cdot \boldsymbol{t} d \Gamma+\int_{\Gamma_{s g}} \boldsymbol{v}^{(k)} \cdot  \boldsymbol{t} d \Gamma+\int_{\Gamma_{s t}} \boldsymbol{v}^{(k)} \cdot \overline{\boldsymbol{t}} d \Gamma \\ & -\int_{\Omega_s} \bm \varepsilon^{(k)}_v : \boldsymbol{\sigma} d \Omega+\int_{\Omega_s} \boldsymbol{v}^{(k)} \cdot \boldsymbol{f} d \Omega
\end{aligned}
\label{eq:residual_elastic_k}
\end{equation}
where $\bm \varepsilon^{(k)}_v = \frac{1}{2} (\nabla \boldsymbol{v}^{(k)} + \nabla \boldsymbol{v}^{(k)T})$ is the symmetric part of $\nabla \boldsymbol{v}^{(k)}$ considering the symmetry of Cauchy stress $\boldsymbol{\sigma}$.

In practice, $N_v = d$ is commonly considered in the study using local variational form~\cite{atluri2000new}. In the following, we take a 2D solid problem as an example, i.e., $d=2$. The set of test functions $\{\bm{v}^{(k)} \}_{k=1}^{2}$ and $\{ \bm \varepsilon^{(k)}_v \}_{k=1}^{2} $ can be assembled in matrices $\bm{w}$ and $\bm{\varepsilon}_w$, respectively, which are
\begin{equation}
\bm w = [v_i^{(j)}]
=\left[\begin{array}{ll} v_{1}^{(1)} & v_{1}^{(2)} \\ v_{2}^{(1)} & v_{2}^{(2)}\end{array}\right], \quad
\bm \varepsilon_w =\left[\begin{array}{ccc} v_{1,1}^{(1)} & v_{2, 2}^{(1)} & v_{1,2 }^{(1)}+v_{2, 1}^{(1)} \\ v_{1,1}^{(2)} & v_{2, 2}^{(2)} & v_{1,2 }^{(2)}+v_{2, 1}^{(2)}
\end{array}\right]^T\
\label{eq:test_function}
\end{equation}
As a result, with invoking the NeuroPU approximation \eqref{eq:nmm} in Eq. \eqref{eq:residual_elastic_k}, the matrix form of the local variational residual at $\Omega_s$ is formulated as
\begin{equation}
\begin{aligned}
\bm{\mathcal{R}}_s^h & =\int_{L_s} \boldsymbol{w}^T  \hat{\boldsymbol{t}}^h d \Gamma+\int_{\Gamma_{s g}} \boldsymbol{w}^T \hat{\boldsymbol{t}}^h d \Gamma+\int_{\Gamma_{s t}} \boldsymbol{w}^T \overline{\boldsymbol{t}} d \Gamma \\ & -\int_{\Omega_s} \bm \varepsilon_w^T \hat{\boldsymbol{\sigma}}^h d \Omega+\int_{\Omega_s} \boldsymbol{w}^T \boldsymbol{f} d \Omega
\end{aligned}
\label{eq:residual_elastic}
\end{equation}
Note that here $\bm{\mathcal{R}}_s^h$ consists of $N_v = 2$ independent equations, but it is straightforward to extend the formulation to the case of $N_v >d$ by using more independent test functions in constructing the local variational residuals.

Furthermore, we can simplify the set of test functions by designing isotropic $\bm w = [v_i^{(j)}] = v \delta_{i j}$ with $v(\bm x)$ being a chosen function defined on $\Omega_s$. Therefore, Eq. \eqref{eq:test_function} is further reduced to
\begin{equation}
\bm w=\left[\begin{array}{ll}v & 0 \\ 0 & v\end{array}\right], \quad \bm \varepsilon_w=\left[\begin{array}{ccc}v_{, 1} & 0 & v_{, 2} \\ 0 & v_{, 2} & v_{, 1}\end{array}\right]^T\
\label{eq:test_simple}
\end{equation}
Unless otherwise stated, the setting of test functions in Eq. \eqref{eq:test_simple} is employed in V-NIM across the present study. The matrix forms of other variables can be referred to \ref{sec:app_A}. 

\emph{Remark 4.1.} It is noted that the trial shape functions $\Psi_I(\bm x)$ in the NeuroPU approximation and the test function $v(\bm x)$ can be chosen from different function spaces, leading to the Petrov-Galerkin method \cite{hughes1982theoretical, bottasso2002discontinuous, atluri1998new}. 
This flexibility allows the test functions defined on \textit{overlapping} subdomains, and the test functions are not required to vanish on the boundary where EBCs are specified (more discussion will be shown in Section \ref{sec:EBC}). 

\emph{Remark 4.2.} We also note that the natural boundary condition (NBC) on $\Gamma_{st}$ are consistently imposed in V-NIM as its boundary integral terms are involved in the local variational form \eqref{eq:local_weak} and the associated residual \eqref{eq:residual_elastic}. Due to the local consistency, we argue that V-NIM can provide a more accurate and stable approximation compared to the weakly imposition of NBC via the penalty method, such as PINN~\cite{raissi2019physics,rao2021physics} and S-NIM \eqref{eq:loss_snim}.

\emph{Remark 4.3.} V-NIM allows the local weak form \eqref{eq:local_weak} (or the local residual form \eqref{eq:residual_elastic}) to be constructed locally with Galerkin consistency.
Given this unique feature, the proposed V-NIM is a truly meshfree framework, distinct from other global variational form-based methods \cite{khodayi2020varnet,kharazmi2021hp,dong2023deepfem, samaniego2020energy, yu2018deep},
where the loss function is constructed by using globally defined test functions or conforming background integration cells over the entire domain. 

\subsubsection{Size of local subdomain}\label{sec:subdomain}
The selection of appropriate subdomains is critical as the local weak formulation is built upon and computed over the set of subdomains $\Omega_s$. While these subdomains can take on arbitrary shapes, in practice, common choices for a local subdomain $\Omega_s$ include a sphere (in the 3D case) or a circle (in the 2D case), as well as a cube (in the 3D case) or a rectangle (in the 2D case)~\cite{atluri1998new,atluri2000new, han2004meshless}.
In this study, we select the square shape with a side length of $2r$ for subdomains. Here, we define $r = \bar r h$, where $h$ represents the characteristic nodal distance, and $\bar r$ serves as a normalized size constant. To ensure appropriate overlaps among subdomains, we adopt $0.5<\bar r<1.5$ for the subdomain size.

\subsubsection{Design of test functions}\label{sec:testfunction}
As indicated before, the local weak form permits the flexibility of defining arbitrary test functions over the local subdomains, including piece-wise polynomials, Legendre polynomials, and radial basis functions. 
In practice, we can customize special test functions on the subdomains to enable distinct properties and simplify the form of local variational residuals \eqref{eq:residual_elastic}. In the following, we will introduce two special properties that can be embedded in the test functions defined on a square subdomain (refer to Section \ref{sec:subdomain}).

\emph{1) Test functions vanishing on local boundary}.
In order to eliminate the boundary integral terms on the part of local boundary $L_s$ (see Figure \ref{fig:NIM}b), we define a test function $v$ that vanishes on $L_s$, i.e., $v (\bm x) = 0$ when $\bm x \in L_s$. To this end, a B-spline function or Gaussian function can be employed. By incorporating these test functions into the local variational residual in Eq. \eqref{eq:residual_elastic}, the boundary integral terms over $L_s$ will be eliminated, and Eq. \eqref{eq:residual_elastic} can be recast as
\begin{equation}
\bm{\mathcal{R}}_s^h = \int_{\Gamma_{s g}} \boldsymbol{w}^T \hat{\boldsymbol{t}}^h d \Gamma+\int_{\Gamma_{s t}} \boldsymbol{w}^T \overline{\boldsymbol{t}} d \Gamma  -\int_{\Omega_s} \bm \varepsilon_w^T \hat{\bm{\sigma}}^h d \Omega +\int_{\Omega_s} \bm{w}^T \boldsymbol{f} d \Omega
\label{eq:residual_elastic_1}
\end{equation}
where $\bm w$ and $\bm \varepsilon_w$ are defined in Eq. \eqref{eq:test_simple}.

\emph{2) Heaviside step function}.
The cumbersome computation of the domain integral in Eq. \eqref{eq:residual_elastic} can be fully circumvented if we adopt the Heaviside step function as the test function, namely,
    \begin{equation}
    v(\boldsymbol{x})= \begin{cases}0 & \boldsymbol{x} \notin \left(\Omega_s \cup L_s\right) \\ 1 & \boldsymbol{x} \in \left(\Omega_s \cup L_s\right)\end{cases}
    \label{eq:heavi}
    \end{equation}
Introducing Eq. \eqref{eq:heavi} into Eqs. \eqref{eq:residual_elastic} and \eqref{eq:test_simple} leads to
    \begin{equation}
    \bm{\mathcal{R}}_s^h=\int_{L_s} \hat{\boldsymbol{t}}^h d \Gamma+\int_{\Gamma_{s g}} \hat{\boldsymbol{t}}^h d \Gamma+\int_{\Gamma_{s t}} \overline{\boldsymbol{t}} d \Gamma+\int_{\Omega_s} \boldsymbol{f} d \Omega
    \label{eq:residual_elastic_2}
    \end{equation}
    where the domain integral term containing the derivative of $v (\bm x)$ is canceled due to the property of Heaviside step function.

It is also worthwhile pointing out that if Dirac’s Delta function $\delta(\bm x - \bm x_s) $, where $\bm x_s$ is the center of subdomain $\Omega_s$, is adopted for test functions, the local variational form \eqref{eq:local_weak} that V-NIM is based on will be degenerated to the strong formulation so that the S-NIM method is restored. In this case, the centers of subdomains $\{\bm x_s\}_{s=1}^{N_{\mathcal{T}}}$ are considered as the sample points $\{\bm x_i\}_{i=1}^{N_f}$ in S-NIM, refer to Figure \ref{fig:NIM}.

\subsubsection{Treatment of essential boundary conditions}\label{sec:EBC}
In the field of physics-informed machine learning, it remains difficult to exactly impose the boundary conditions including EBC and NBC. 
As suggested in Remark 4.2, NBC is consistently considered in the local Petrov-Galerkin formulation in V-NIM, so we are only concerned with the enforcement of EBC in this subsection.

The penalty method has been widely used in traditional numerical methods, we borrow this idea by adding additional terms on the local variational loss $\bm{\mathcal{R}}_s^h$
    \begin{equation}
\bm{\mathcal{R}}_s^{h,g}= \bm{\mathcal{R}}_s^h + \alpha \int_{\Gamma_{s g}} \bm w^T (\hat{\boldsymbol{u}}^h - \bar{\bm u}) d \Gamma
\label{eq:penalty}
    \end{equation}
where $\alpha$ is a penalty parameter. 

It is noted that since PU shape functions are adopted to construct the approximation \eqref{eq:nmm}, various boundary enforcement methods proposed in Meshfree community~\cite{chen2017meshfree} can be readily applied to the V-NIM method, such as Lagrangian method, singular kernel method, or Nitche’s method. Here, we only consider the penalty method for the sake of simplicity and consistency in comparison with the S-NIM and PINN methods.

\subsubsection{Loss function of V-NIM}
Integrating the local variational residuals over the set of subdomains $ \{\Omega_s\}_{s=1}^{N_\mathcal{T}}$ and considering the set of system parameters, the total loss of the V-NIM is written as:
\begin{equation}
\begin{aligned}
\mathcal{L}^V(\bm \theta)&=\frac{1}{N_{\mu}N_\mathcal{T}} \sum_{j=1}^{N_{\mu}}\sum_{s=1}^{N_\mathcal{T}}\left\|\bm{\mathcal{R}}_s^{h,g}(\bm \mu_j) \right\|^2 \\ &= \frac{1}{N_{\mu}N_\mathcal{T}}\sum_{j=1}^{N_{\mu}}\sum_{s=1}^{N_\mathcal{T}}\left\|\bm{\mathcal{R}}_s^h + \alpha \int_{\Gamma_{s g}} \bm w^T (\hat{\boldsymbol{u}}^h - \bar{\bm u}) d \Gamma \right\|^2
\label{eq:loss_v}
\end{aligned}
\end{equation}
where some examples of the local variational residual $\bm{\mathcal{R}}_s^h$ are provided in Section \ref{sec:testfunction}.
Again, $\hat{\bm{u}}^h(\bm x,\bm \mu; \bm \theta) = \sum_{I \in \mathcal{S}_x} \Psi_I(\bm{x}) \hat{\bm d}_I(\bm \mu; \bm \theta)$ adopts the NeuroPU approximation, where $\bm \theta$ are the trainable parameters of the neural networks.

As reported in Remark 4.3, it can be seen that the loss $\mathcal{L}^V$ for V-NIM simply relies on the summation of local variational residuals instead of a full integral form over the whole domain to enforce equilibrium~\cite{khodayi2020varnet,samaniego2020energy, yu2018deep,berrone2022variational}.
Thanks to this collocation-like construction, each local residual can be minimized separately.
On the other hand, compared to other domain decomposition-based DNN methods, such as \textit{hp}-VPINN~\cite{kharazmi2021hp} or local extreme learning machines~\cite{dong2021local}, the proposed V-NIM doesn't require conforming subdomains to formulate the local residuals, which offers a great flexibility in constructing the loss function. 
Overall, the local feature embedded in V-NIM enables the employment of an efficient and scalable (mini-)batch training procedure. The enhanced computational efficiency and accuracy will be highlighted in the following numerical experiments. 

\section{Solution Procedures}\label{sec:procedure}

A summary of the numerical procedures for the proposed S-NIM and V-NIM methods is provided in this section.

\subsection{Numerical procedures}

\subsection*{Implementation of the S-NIM method}
\begin{enumerate}
    \item[1.] Define the following parameters.
\begin{enumerate}
    \item[1.a.] The sets of meshfree nodes $\{\bm{x}_I\}_{I=1}^{N_h}$ and residual points $\mathcal{S}_f$  distributed over $\Omega$. The sets of sample points $\mathcal{S}_t$, $\mathcal{S}_g$ distributed on $\Gamma_t$ and $\Gamma_g$, respectively.
    \item[1.b.] The parameters for NeuroPU approximation (Figure \ref{fig:structure}) including the neural network architecture, the order of basis function $p$, the support size $a$, and the type of shape functions.
\end{enumerate}

    \item[2.] Calculate and store the trial shape functions $\{\Psi_I\}_{I=1}^{N_h}$ associated with nodes $\{\bm{x}_I\}_{I=1}^{N_h}$, and initialize the nodal coefficient network $\hat{\bm d} (\bm \mu;\bm \theta) = \mathcal{N}_{\theta}(\bm \mu)$.
    \item[3.] Construct the loss function based on Eq. \eqref{eq:loss_snim}. 
    \item[4.] Define the optimizer and minimize the loss function until convergence.
    \item[5.] Output the trained network $\hat{\bm{d}}(\bm{\mu};\bm \theta^*)$.
\end{enumerate}

\subsection*{Implementation of V-NIM method}
\begin{enumerate}
    \item[1.] Define the following parameters.
\begin{enumerate}
    \item[1.a.] The set of meshfree nodes $\{\bm{x}_I\}_{I=1}^{N_h}$ and the set of center points of subdomains $ \{\bm{x}_s\}_{s=1}^{N_\mathcal{T}}$ distributed over $\Omega$.
    \item[1.b.] The parameters for NeuroPU approximation (Figure \ref{fig:structure}) including the neural network architecture, the order of basis function $p$, the support size $a$, and the type of shape functions.
    \item[1.c.] The parameters for local variational form including the size of subdomains $r$ and the type of test functions.
\end{enumerate}

    \item[2.] Calculate and store the trial shape functions $\{\Psi_I\}_{I=1}^{N_h}$ associated with nodes $\{\bm{x}_I\}_{I=1}^{N_h}$, and initialize the nodal coefficient network $\hat{\bm d} (\bm \mu;\bm \theta) = \mathcal{N}_{\theta}(\bm \mu)$.
    \item[3.] Construct the loss function based on Eq. \eqref{eq:loss_v} by using the Gauss quadrature points.
    \begin{enumerate}
    \item[3.a.] By introducing the quadrature rules, the discrete forms of Eq. \eqref{eq:residual_elastic_1} and Eq. \eqref{eq:residual_elastic_2} are, respectively, given as follows

\begin{equation}
\begin{aligned}
\bm{\mathcal{R}}_s^h  & = \int_{\Gamma_{s g}} \boldsymbol{w}^T \boldsymbol{t}^h d \Gamma+\int_{\Gamma_{s t}} \boldsymbol{w}^T \overline{\boldsymbol{t}} d \Gamma  -\int_{\Omega_s} \bm \varepsilon_w^T \boldsymbol{\sigma}^h d \Omega+\int_{\Omega_s} \boldsymbol{w}^T \boldsymbol{f} d \Omega \\ 
& =\sum_{E=1}^{N_L}\left\{J_{\Gamma_{s g}^E} \sum_{B=1}^{N_B} \boldsymbol{w}^T\left(\boldsymbol{x}_B\right) \boldsymbol{t}^h\left(\boldsymbol{x}_B\right) \omega_B\right\} 
\\& +\sum_{E=1}^{N_L}\left\{J_{\Gamma_{s t}^E} \sum_{B=1}^{N_B} \boldsymbol{w}^T\left(\boldsymbol{x}_B\right) \overline{\boldsymbol{t}}\left(\boldsymbol{x}_B\right) \omega_B\right\} 
\\ & -\sum_{E_x=1}^{N_E^x} \sum_{E_y=1}^{N_E^y} \left\{J_{\Omega_s^{\left(E_x, E_y\right)}}\sum_{G=1}^{N_G}\bm \varepsilon_w^T\left(\boldsymbol{x}_G\right) \boldsymbol{\sigma}^h\left(\boldsymbol{x}_G\right)\omega_G\right\}
\\ & +\sum_{E_x=1}^{N_E^x} \sum_{E_y=1}^{N_E^y} \left\{J_{\Omega_s^{\left(E_x, E_y\right)}}\sum_{G=1}^{N_G}\boldsymbol{w}^T\left(\boldsymbol{x}_G\right) \boldsymbol{f}\left(\boldsymbol{x}_G\right) \omega_G\right\}
\end{aligned}
\label{eq:dis_1}
\end{equation}
and
\begin{equation}
\begin{aligned}
\bm{\mathcal{R}}_s^h & =\int_{L_s} \boldsymbol{t}^h d \Gamma+\int_{\Gamma_{s g}} \boldsymbol{t}^h d \Gamma+\int_{\Gamma_{s t}} \overline{\boldsymbol{t}} d \Gamma+\int_{\Omega_s} \boldsymbol{f} d \Omega \\ & =\sum_{E=1}^{N_L}[J_{L_s^E} \sum_{B=1}^{N_B} \boldsymbol{t}^h\left(\boldsymbol{x}_B\right) \omega_B]+\sum_{E=1}^{N_L}[J_{\Gamma_{s g}^E} \sum_{B=1}^{N_B} \overline{\boldsymbol{t}}\left(\boldsymbol{x}_B\right) \omega_B] \\ & +\sum_{E=1}^{N_L}[J_{\Gamma_{s t}^E} \sum_{B=1}^{N_B} \overline{\boldsymbol{t}}\left(\boldsymbol{x}_B\right) \omega_B]
+\sum_{E_x=1}^{N_E^x} \sum_{E_y=1}^{N_E^y}[J_{\Omega_s^{\left(E_x, E_y\right)}} \sum_{G=1}^{N_G} \boldsymbol{f}\left(\boldsymbol{x}_G\right) \omega_G]
\end{aligned}
\label{eq:dis_2}
\end{equation}
where $\bm x_B$'s and $\omega_B$'s are the locations and weights of quadrature points for boundary integrals, while $\bm x_G$'s and $\omega_G$'s correspond to those for domain integrals. $J$ stands for the corresponding Jacobian. $N_L$ represents the number of segments for the boundary integral. $N_E^x$ and $N_E^y$ are the number of segments for the domain integral along $x$ and $y$ directions, respectively, when considering a 2D rectangle subdomain.
\item[3.b.] The essential boundary terms in \eqref{eq:loss_v} will be discretized in the similar way shown in Eqs. \eqref{eq:dis_1} and \eqref{eq:dis_2}.
\end{enumerate}
\item[4.] Define the optimizer and minimize the loss function until convergence.
\item[5.] Output the trained network $\hat{\bm{d}}(\bm{\mu};\bm \theta^*)$.
\end{enumerate}

Once the trained network $\hat{\bm{d}}(\bm{\mu};\bm \theta^*)$ is given, the solution field is obtained by using the NeuroPU approximation: $\hat{\bm{u}}^h(\bm{x},\bm{\mu}) = \sum_{I \in \mathcal{S}_x} \Psi_I(\bm{x}) \hat{\bm{d}}(\bm{\mu};\bm \theta^*)$.

\subsection{Summary: S-NIM \& V-NIM}
Table \ref{tab:formulation} provides an overview of the proposed S-NIM and V-NIM solvers.
While we only consider two different test functions, the framework can be easily extended to other types of functions depending on applications of interest (see Remark 4.1). We also refer to Table 1 in \cite{kharazmi2021hp}, which provides a summary of test function and trial function for various variations and energy-based PINN approaches. 
In the next section, we will compare the performance of these proposed methods through a series of numerical examples.
\begin{table}[htb]
\centering
\small
\begin{tabular}{|c|c|c|c|}
  \hline
  Methods & S-NIM & \multicolumn{2}{c|}{V-NIM} \\
  \hline
  Trial function/solution  & \multicolumn{3}{c|}{NeuroPU approximation~\eqref{eq:nmm}}  \\
  \hline
  Test function & $\delta(\bm x - \bm x_s)$ & Heaviside step & Cubic B-spline  \\
  \hline
  Loss/Residual & Strong form \eqref{eq:loss_snim} & \multicolumn{2}{c|}{Local variational form \eqref{eq:loss_v}} \\
  \hline
   Integration & N/A & N/A & Gauss quadrature  \\
  \hline
\end{tabular}
\caption{Summary of the two proposed NIM methods: S-NIM and V-NIM.}
\label{tab:formulation}
\end{table}

The proposed NIM framework is implemented based on TensorFlow 1.14 in Python 3.9 for leveraging its built-in automatic differentiation capacity. The computations are executed on a single NVIDIA A100 graphics processing unit (GPU).

\section{Numerical Results: Static Problem}\label{sec:result}
In this section,
we will examine the performance of the proposed S-NIM and V-NIM methods on various numerical examples. 
The approximation and convergence properties of the NeuroPU approach and the features of the local formulation will be investigated using the static scalar (Poisson's equation) and vector (elasticity) problems. The superior performance of V-NIM based on the local weak formulation will be highlighted.
Moreover, we will demonstrate that the NIM framework can be used for surrogate modeling of a parameterized PDE with a desirable extrapolative capacity. 
In Section \ref{sec:result_ade}, the proposed method will be further validated on a time-dependent problem, the advection-diffusion equation, where time is considered as the model parameter. 
The solutions obtained by the PINN and \textit{hp}-VPINN methods \cite{kharazmi2021hp} are also provided for comparison to underscore the advantages of the NIM solver in both approximation accuracy and computational efficiency.

Unless stated otherwise, the construction of the NeuroPU approximation \eqref{eq:nmm} in the NIM methods employ quadratic meshfree shape functions, i.e., $p=2$ in \eqref{eq:basis}, with the normalized support size $\bar a =2.5$. 
To simplify notation, a neural network with $l$ hidden layers, each of which contains $m$ neurons, is denoted as $l \times [m]$. The input size of the neural network in NeuroPU approximation is given by the dimension of parameters $\bm \mu \in \mathbb{R}^c$, and the output size is $N_h$. 

As shown in Table \ref{tab:formulation}, two different test functions are examined in V-NIM. 
To distinguish the employment of the Heaviside function and cubic B-spline function as the test functions, we denote the resultant V-NIM solvers as $\text{V-NIM/h}$ and $\text{V-NIM/c}$, respectively. 
As domain integral is required in V-NIM, we let each subdomain be uniformly divided into $4\times4$ segments in 2D case (or 4 segments in 1D case), where $5$ Gauss quadrature points per direction are used for the segment integration. 

The initial weights and bias of neural networks in the NIM framework are initialized using the Xavier scheme \cite{glorot2010understanding}.
For the penalty parameters $\alpha$ adopted for the S-NIM \eqref{eq:loss_snim} and V-NIM \eqref{eq:loss_v} methods, we determine the optimal one from the heuristic tests on the values $[1, 10, 10^2, 10^3]$.
In order to conduct a fair comparison, we consider the same training scheme for different methods to evaluate their training efficiency and accuracy.
The Adam optimizer with a learning rate of $0.001$ is used by default unless stated otherwise.

\subsection{2D Poisson's equation}\label{sec:pro_poi}
To demonstrate the accuracy and convergence properties of the NIM solvers, we first consider a two-dimensional Poisson's equation:
\begin{equation}
\begin{cases}\nabla^2 u(x, y)=f(x, y) & \text { in } \Omega:[-1,1] \times[-1,1] \\ u(x, y)=\bar{u} & \text { on } \Gamma_g: x= \pm 1 \text { or } y= \pm 1\end{cases}
\label{eq:2D_poisson}
\end{equation}
where $f(x,y)$ and $\bar u$ are the prescribed force term and the essential boundary condition (EBC) on $\Gamma_g$, respectively.
Let the analytical solution~\cite{kharazmi2019variational,kharazmi2021hp} be 
\begin{equation}
u^{\text {exact }}(x, y)=(0.1 \sin (2 \pi x)+\tanh (10 x)) \times \sin (2 \pi y)
\label{eq:analytical}
\end{equation}
The corresponding $f(x,y)$ and $\bar u$ can be obtained by substituting the exact solution \eqref{eq:analytical} in \eqref{eq:2D_poisson}.
In this static problem, since the parameter coefficient $\bm \mu$ is considered a constant, it is dropped in Eq. \eqref{eq:2D_poisson} for simplicity.
As a result, a shallow neural network (see Table \ref{tab:setting_poi}) is selected for the NeuroPU approximation. 

\begin{figure}[htb]
	\centering
 \includegraphics[angle=0,width=0.7\textwidth]{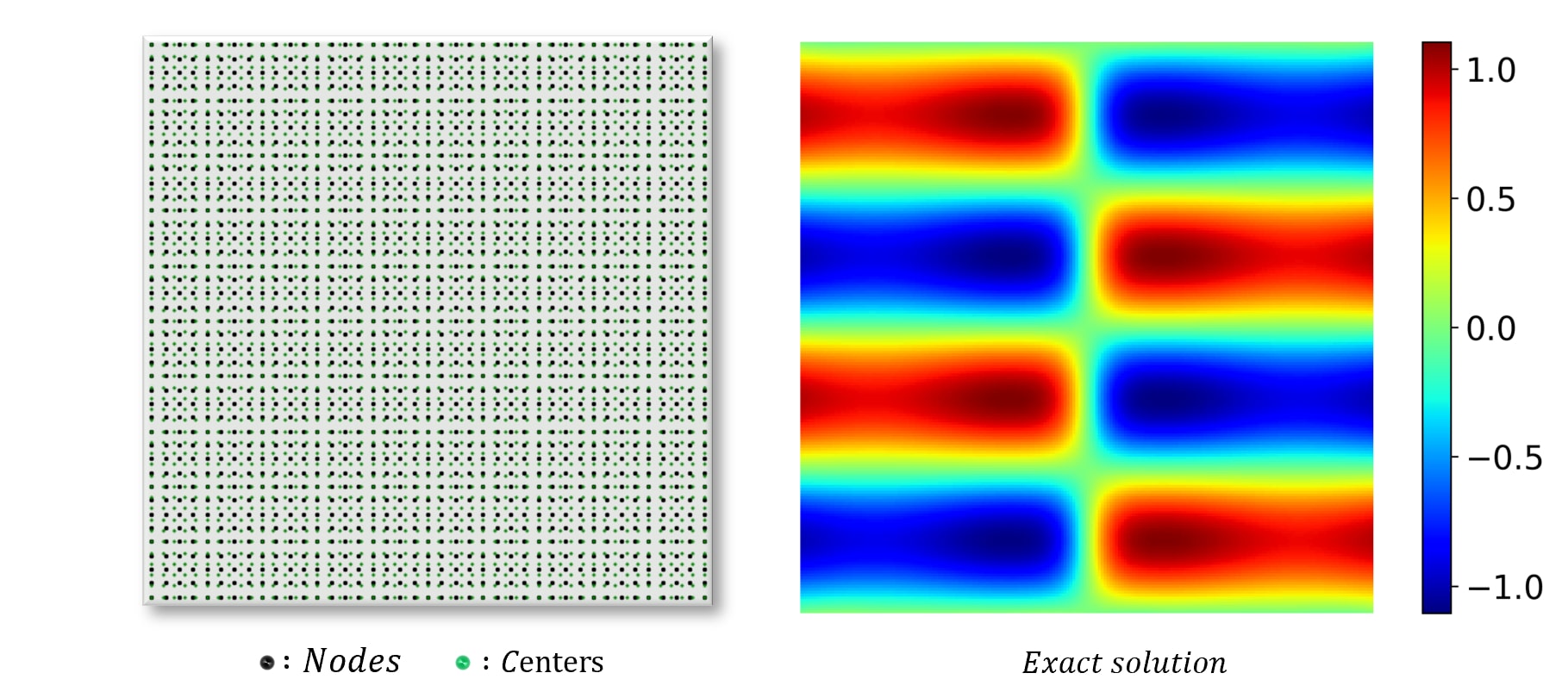}
	\caption{Left: The discretization for NIM, where the nodes are used to define the nodal shape function and the center points represent the sample points for S-NIM and centers of subdomains for V-NIM; Right: The reference solution for the 2D Poisson's problem.}
    \label{fig:poi_pro}
\end{figure}

\begin{table}[htb]
\centering
\small
\begin{tabular}{|c|c|c|c|}
  \hline
Methods & PINN & S-NIM & V-NIM\\
  \hline
 Neural network &  4$\times$[40]  &1$\times$[10] & 1$\times$[10] \\
 \hline
  $N_h$ & N/A &  \multicolumn{2}{c|}{1681}   \\
  \hline
  Subdomain size $\bar r$ &  N/A & N/A &  1.5 \\
  \hline
  $N_f$ or $N_{\mathcal{T}}$ & 10400  & 2601 &  2601 \\
  \hline
  $\alpha$ & 100 & 1000 & 100\\
  \hline
\end{tabular}
\caption{Hyperparameters of PINN, S-NIM, and V-NIM for 2D Poisson problem.}
\label{tab:setting_poi}
\end{table}

\subsubsection{Effect of NeuroPU approximation}\label{sec:Effect}
The effect of NeuroPU approximation with different orders of basis on the NIM solutions is investigated in this subsection.
As shown in Figure \ref{fig:poi_pro}(left), a uniform meshfree discretization of $N_h=1681$ nodes is adopted for the NeuroPU approximation \eqref{eq:nmm}, whereas a set of 2601 uniformly distributed residual points (or subdomains) is used for S-NIM (or V-NIM) method. 
The comparison of the V-NIM/h and V-NIM/c methods with the same normalized size $\bar r=1.5$ of subdomains is also provided. 
The network architecture of hidden layers for the NIM methods is $1 \times [10]$, and the output layer has $N_h = 1681$ neurons. 
We set the number of Adam epochs to 50,000. All the hyperparameters are provided in Table \ref{tab:setting_poi}. 

\begin{figure}[htb]
	\centering
	\includegraphics[angle=0,width=0.8\textwidth]{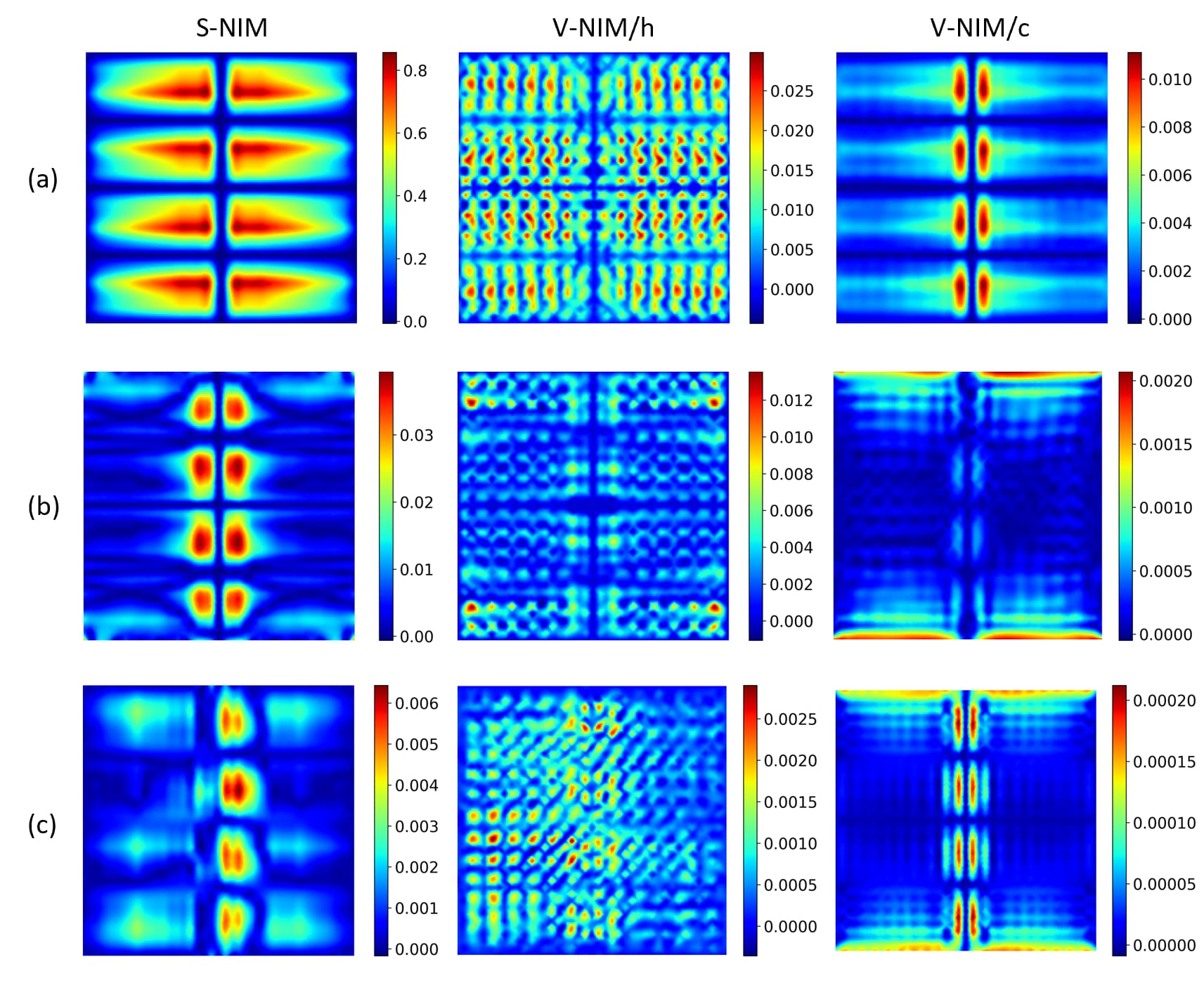}
	\caption{The comparisons of point-wise displacement errors obtained by S-NIM, V-NIM/h and V-NIM/c using different orders of NeuroPU shape functions: (a) Linear; (b) Quadratic; (c) Cubic. The reference solution is provided in the right panel of Figure \ref{fig:poi_pro}.}
    \label{fig:three}
\end{figure}
 
Figure \ref{fig:three} presents the point-wise error comparison among S-NIM, V-NIM/h and V-NIM/c using different orders of basis ($p=1,2,3$ with $\bar a=1.5,2.5$ and $3.5$) for the NeuroPU shape functions. 
As the order of basis increases, the approximation errors of all these NIM methods reduce significantly. For example, the maximum point-wise error in S-NIM decreases from around $0.85$ with a linear basis to $6 \times 10^{-3}$ with a cubic basis, whereas that of V-NIM/h from $1 \times 10^{-2}$ to $2 \times 10^{-4}$. 
This highlights the flexibility of using high-order approximation to achieve desirable accuracy in the NIM solution even with a relatively small network, which is also evidenced by the convergence study (Figure \ref{fig:converge_poission}). 
It is important to note that the diminished accuracy in the S-NIM solution using linear basis, as shown in Figure \ref{fig:three}a, stems from the involvement of 2nd order derivatives in the strong-form PDE residuals, which results in the linear RK shape functions lacking sufficient approximation capacity. 

Comparing the strong form and local weak form-based NIMs, it is observed that both V-NIM/h and V-NIM/c exhibit higher accuracy than S-NIM, especially in the case of V-NIM/c, which surpasses S-NIM by more than an error order of magnitude.
This mainly attributes to the following properties provided by the local variational formulation. First, the residuals of V-NIM~\eqref{eq:residual_elastic} involve lower order of derivatives than S-NIM (and PINN, refer to Section \ref{sec:nim_pinn}), which reduces the NeuroPU approximation error related to derivatives; Second, the proposed variational approach yields a consistent residual form to satisfy the local equilibrium given its loss is adequately minimized; Third, due to the compact support of test functions, the nonlocal information over a subdomain can be incorporated in the residuals, whereas S-NIM only captures the misfit information locally at the sample points.

On the other hand, we observe that V-NIM using the cubic B-spline test function (V-NIM/c) yields superior performance compared to V-NIM with a Heaviside step function (V-NIM/h). 
This is understandable because the test functions with higher order continuity will improve the accuracy and stability in the local weak form, at the expense of obtaining a more complicated residual form (see Eqs. \eqref{eq:residual_elastic_1} and \eqref{eq:residual_elastic_2}).

\subsubsection{NIM versus PINN}\label{sec:nim_pinn}
To highlight the superior accuracy and efficiency of the proposed hybrid framework, we compare the solutions obtained by the standard PINN model and the S-NIM and V-NIM model, in which the cubic trial shape function ($p=3$) with a normalized support size of $\bar{a} = 3.5$ is adopted for NeuroPU approximation.
The hyperparameters used for S-NIM and V-NIM remain the same as Table \ref{tab:setting_poi}.
To ensure sufficient accuracy in the PINN solution, we utilize $N_f = 10,400$ uniformly distributed residual points, whereas only $N_{\mathcal{T}}= 2,601$ subdomains (or $N_f= 2,601$ sample points) are considered in V-NIM (or S-NIM). 
A neural network with hidden layers $4 \times [40]$ is adopted for the PINN method.

\begin{figure}[htb]
	\centering
    \includegraphics[angle=0,width=1.0\textwidth]{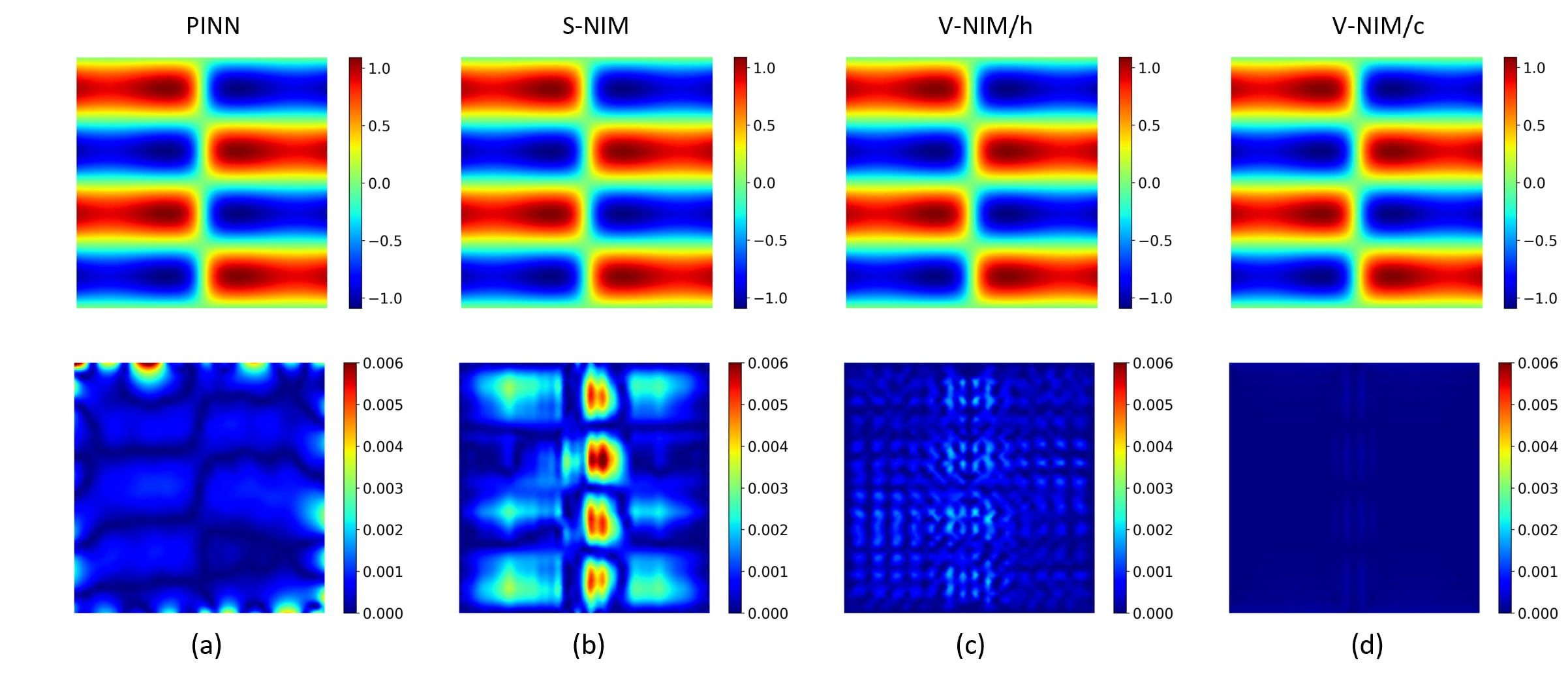}
	\caption{Comparison of the approximated displacement (upper row) and the absolute point-wise errors (lower row) obtained by (a) PINN, (b) S-NIM, (c) V-NIM/h and (d) V-NIM/c. The cubic NeuroPU shape function is adopted for all NIM methods.}
    \label{fig:poi_comp_nim}
\end{figure}

Figure \ref{fig:poi_comp_nim} shows the comparison of the approximate solutions obtained by different NIMs and PINN, and their corresponding absolute point-wise errors. The maximum errors in solution approximation for all cases are less $0.01$. Compared to PINN, the results indicate that S-NIM generally provides a slightly improved approximation on the edge regions, albeit with a slightly larger error over the higher-gradient areas. Nevertheless, V-NIM yields the most accurate results, indicating the preferable accuracy of the proposed V-NIM methods over the PINN method. 

Furthermore, the comparison of corresponding mean absolute errors ($MAE$) and training costs are listed in Table \ref{tab:poi_results}.
The $MAE$ of S-NIM using 2601 residual points is $1.11 \times 10^{-3}$, which showcases a slight degradation over the PINN method with $MAE = 6.32 \times 10^{-4}$. However, this is because S-NIM only utilizes one fourth of sample points compared to PINN. In terms of training cost, PINN takes approximately 10.65s for every 1000 epochs, approximately 8 times slower than S-NIM, as illustrated in Table \ref{tab:poi_results}. Furthermore, in order to demonstrate the superior ability of S-VIM over PINN method, 
we adopt an increased number of residual points ($N_f = 10,000$) to train the S-NIM model, denoted as \textit{S-NIM2} in Table \ref{tab:poi_results}.
The corresponding $MAE$ error is substantially reduced to $2.19 \times 10^{-4}$ from $1.11 \times 10^{-3}$.
The training time for S-NIM2 becomes $2.01s/1000$ epochs, which is 1.4 times longer than the S-NIM case with $N_f = 2,601$.
Overall, this refined S-NIM solution yields 3 times higher accuracy and 5 times higher efficiency compared with the PINN method,
demonstrating the enhancement by introducing the NeuroPU approximation in S-VIM.

The V-NIM methods further improve the performance, outperforming PINN by approximately 1.5 times and 8 times in terms of accuracy when using V-NIM/h and V-NIM/c, respectively.
This remarkably higher accuracy is attributed to the local variational form as we described in Section \ref{sec:v-nim}. Consistent with the observation in Section \ref{sec:Effect}, the employment of smooth test functions (V-NIM/c) leads to remarkably higher accuracy in approximating the solution.

Because the shape functions in NIMs are pre-computed and stored, similar to the approach in FEM, the additional computational cost incurred from using higher-order approximation is, in fact, quite marginal during online training. It is interesting to observe that, owing to the lower order derivatives involved in the weak-form residuals, V-NIM even demonstrates superior training efficiency to S-NIM. Specifically, the training of V-NIM for every 1000 epochs costs approximately $1.20$s, resulting in a 1.2x faster training rate than that of S-NIM. 
The efficiency enhancement becomes more pronounced when compared to PINN, which leads to a 10x speedup.

\begin{table}[htb]
\centering
\small
\begin{tabular}{|c|c|c|c|c|c|}
  \hline
Methods & PINN & S-NIM & S-NIM2 & V-NIM/h & V-NIM/c\\
  \hline
   Epochs& \multicolumn{5}{c|}{50000}  \\
  \hline
  \makecell{Training time: \\ {[s]}/1000 epochs} & 10.65 & 1.46 & 2.01 & 1.21 & 1.20\\
  \hline
  $MAE$ & 6.32e-04 & 1.11e-03 & 2.19e-04 & 4.21e-04 & 7.89e-05\\
  \hline
\end{tabular}
\caption{Training results of PINN, S-NIM, V-NIM/h and V-NIM/c for 2D Poisson's problem, where the corresponding hyperparameters are referred to Table \ref{tab:setting_poi}. For the demonstration of the effect of sample points, we also provide the S-NIM2 solution that is obtained by using $N_f = 10,000$ sample points.}
\label{tab:poi_results}
\end{table}

\subsubsection{Convergence study}\label{sec:poi_conv}
The preceding results show the outstanding performance of the proposed V-NIM framework, resulting from the introduction of NeuroPU and local weak formulation.
In this subsection, we will illustrate the convergence property of V-NIM in relation to meshfree discretization, exploring various combinations of test and trial functions. 
We will consider four distinct discretizations, comprising 121, 441, 961 and 1681 uniformly distributed nodes, respectively. 
Note that we have excluded the discussion of strong-form methods due to their lack of clearly convergent behavior under the same training conditions.

In this test, we adopt 100,000 epochs with a decaying learning rate from $10^{-3}$ to $10^{-5}$ for Adam optimizer and a fixed number of subdomains ($N_{\mathcal{T}}= 2,601$) to ensure a stable solution.
The normalized sizes of support domain are set as $\bar a = 2.5$ for quadratic shape function and $\bar a = 3.5$ for cubic shape function, respectively. The other unstated parameters are referred to Table \ref{tab:setting_poi}.
To properly measure the solution errors, we define the errors in relative $L^2$ norm and relative semi-$H^1$ norm, namely, $e_0$ and $e_1$, as follows:
\begin{equation}
e_0=\frac{\sqrt{\int_{\Omega}\left(u-u^h\right)^2 d \Omega}}{\sqrt{\int_{\Omega} u^2 d \Omega}}, \quad c_1=\frac{\sqrt{\int_{\Omega} \sum_{i=1}^2\left(u_{, i}-u_{, i}^h\right)^2 d \Omega}}{\sqrt{\int_{\Omega} \sum_{i=1}^2 u_{, i}^2 d \Omega}}
\end{equation}
where $u$ and $u^h$ denote the reference and predicted solution, respectively. 

\begin{figure}[htb]
	\centering
    \includegraphics[angle=0,width=1.0\textwidth]{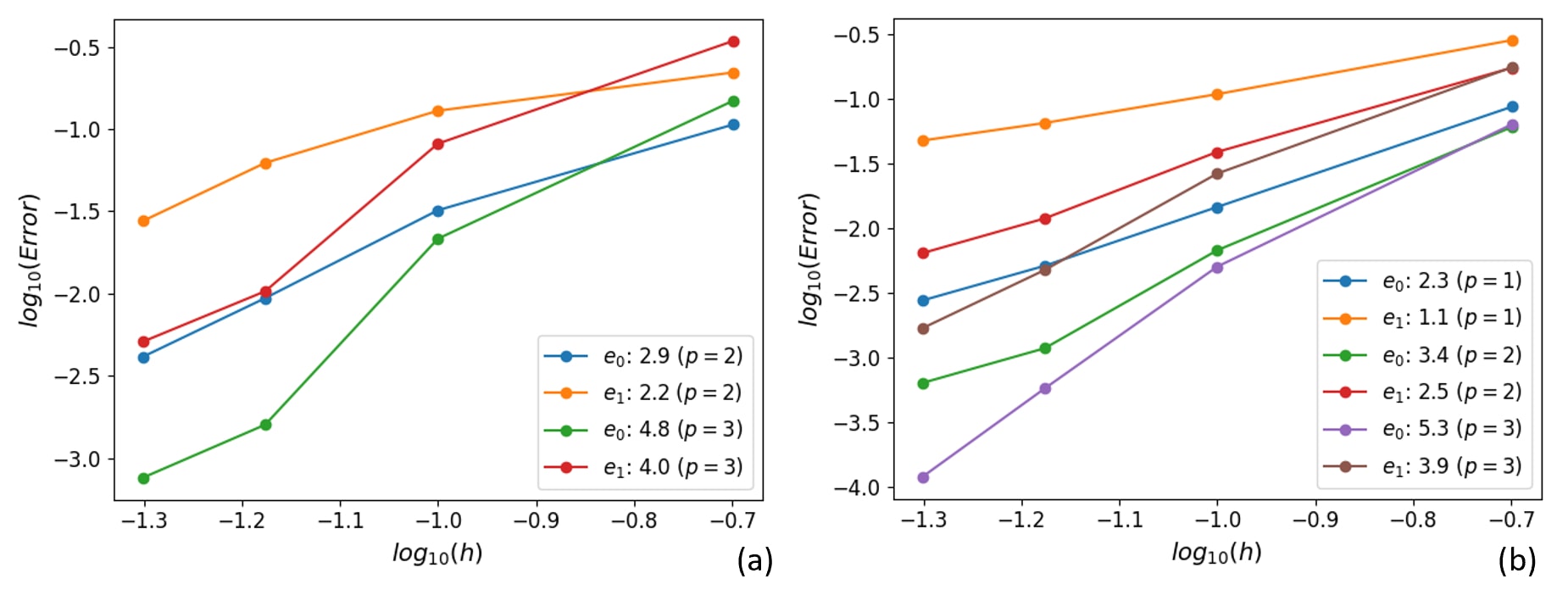}
    \caption{Convergence study of the V-NIM methods using different test functions: (left) V-NIM/h with Heaviside step function; (right) V-NIM/c with cubic B-spline function.
    The convergence rates presented in the legend are determined by calculating the average slope of the last two segments. $p$ denotes the order of basis used for the NeuroPU approximation function in the NIM methods.}
    \label{fig:converge_poission}
\end{figure}

Figure \ref{fig:converge_poission} displays the convergence results of V-NIM/h (left) and V-NIM/c (right) when employing different-order bases ($p = 1,2,3$) in the trial (NeuroPU) shape functions. 
In the case of V-NIM/h, we do not consider the linear shape function ($p = 1$) as the resultant weak form solution becomes unstable due to the low-order continuity of the Heaviside step function.
Overall, the results show that both the $e_0$ and $e_1$  errors of the NIM methods progressively decrease as refining the meshfree discretizations, where the rates of asymptotic convergence are also provided in the figure. 
It is interesting to notice that these convergence rates approximately follow the error estimates derived from the classical FEM or Meshfree approximation~\cite{hughes2012finite,chen2017meshfree}, i.e., $e_0 \sim  O(h^{p+1})$ and $e_1 \sim  O(h^{p})$, where $h$ represents the characteristic nodal distance and $p$ is the order of basis for trial functions. 
For example, the convergence rates of $e_0$ and $e_1$ produced by V-NIM/h, using quadratic ($p = 2$) NeuroPU shape functions, are approximately $r_{e_0}=2.9$ and $r_{e_1}=2.2$, respectively.
On the other hand, when employing V-NIM/c with linear and quadratic NeuroPU shape functions, the convergence rates for $e_0$ are observed to be $r_{e_0}=2.3$ and $3.4$, while those for $e_1$ are $r_{e_1}=1.1$ and $2.5$, respectively. 
Surprisingly, we notice that the cases with cubic shape functions ($p = 3$) yield increased convergence rates in V-NIMs.

It is clearly shown in Figure \ref{fig:converge_poission} that in comparison to V-NIM/h, V-NIM/c tends to offer better stability and attain favorable accuracy due to the high continuity of the test function.

\subsection{Linear elasticity: Defected plate}\label{sec:pro_ela}

\begin{figure}[htb]
	\centering
	\includegraphics[angle=0,width=0.9\textwidth]{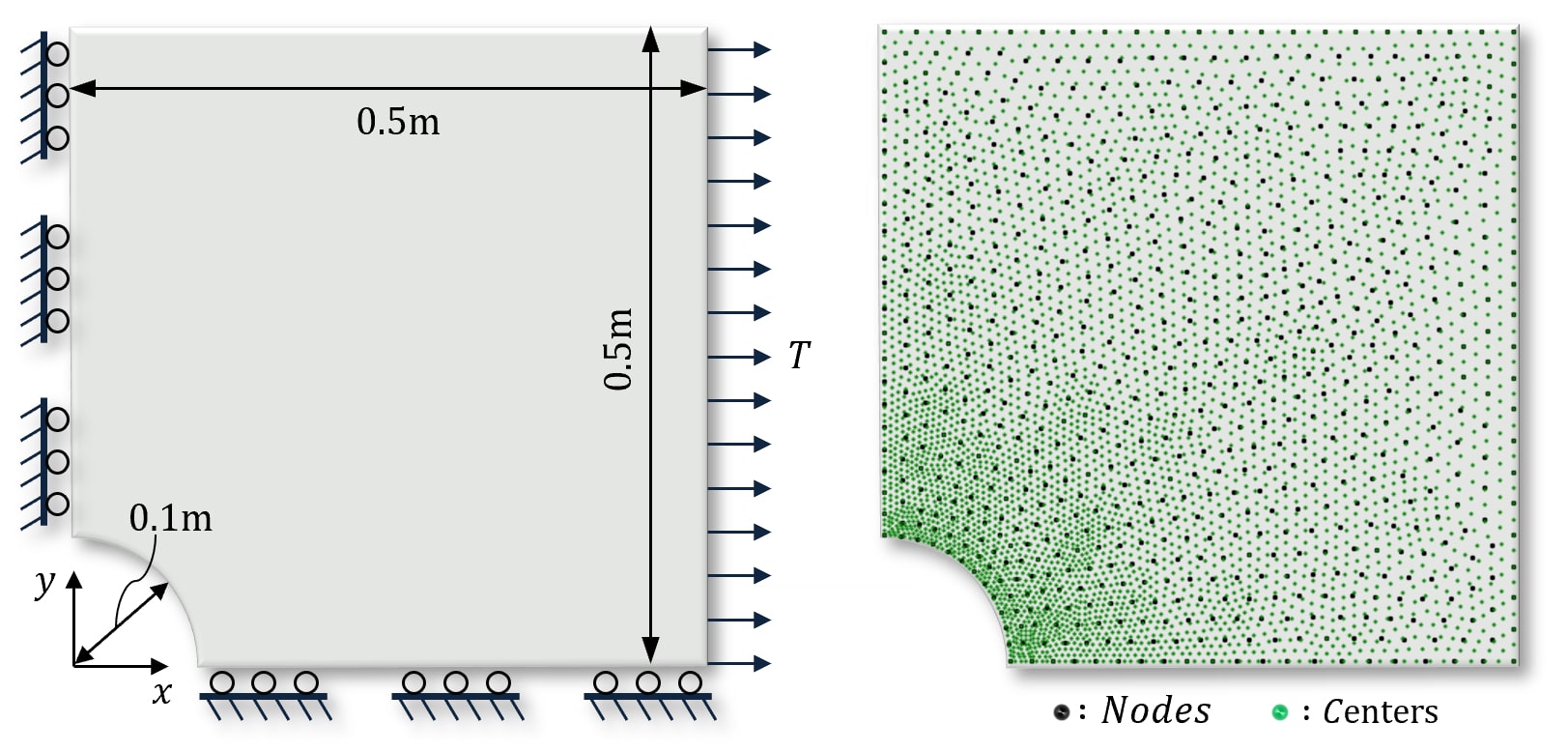}
	\caption{Left: Schematic of defected plate under normal traction; Right: Distribution of nodes (black points) and centers of subdomains (green points).}
    \label{fig:plate}
\end{figure}

In this example, we consider a plane-stress elastic square plate subjected to a uniform normal traction, as shown in Figure \ref{fig:plate}(left).
Due to the symmetry of the problem, only a quarter plate with a length of $0.5$ m is simulated, where the Young’s modulus $E$ and Poisson’s ratio $\nu$ are $20$ \text{MPa} and $ 0.25$, respectively, and the traction imposed on the right side is $T=1$ \text{MPa}.

For demonstration, the V-NIM solver with Heaviside test function (V-NIM/h) is adopted to solve this problem, where the associated normalized size of subdomains is set as $\bar r=1.2$. 
The model discretization with 727 nodes and 4117 subdomains is shown in Figure \ref{fig:plate}(right). 
A neural network with one hidden layer (10 neurons), i.e., $1 \times [10]$, and the quadratic trial shape function with normalized support size of $\bar a=2.4$ are employed to construct the NeuroPU approximation in V-NIM/h. 
As it was reported that the standard PINN performs inefficiently in elasticity problems, the mixed-variable based PINN method developed in \cite{rao2021physics} for elasticity mechanics is adopted here as the baseline for comparison. 
We note that, consistent with the previous examples, the penalty method is still used to impose the boundary conditions for both V-NIM/h and the mixed-variable PINN, in contrast to the hard enforcement of boundary conditions as reported in \cite{rao2021physics}.
The architecture of PINN is set as $5 \times [50]$, and 22000 residual points are fed for training. An Adam optimizer with 150,000 epochs and a decaying learning rate from $10^{-3}$ to $10^{-5}$ is utilized for both PINN and V-NIM/h methods. 
All the hyperparameters of method settings are listed in Table \ref{tab:setting_plate}.

\begin{table}
\centering
\small
\begin{tabular}{|c|c|c|}
  \hline
  Methods & {V-NIM/h} & PINN \\
    \hline
  Neural networks& {1$\times$[10]} & {5 $\times$ [50]} \\
   \hline
    $N_h$ & {727} & {N/A}\\
 \hline
  Subdomain size $\bar r$ & {1.2} & {N/A}\\
 \hline
  Support size $\bar a$ & {2.4} & {N/A}\\
  \hline
  $N_f$ or $N_{\mathcal{T}}$  & {4117}  & {22000}\\
  \hline
  $\alpha$ & 100 & 100  \\
\hline
  Segments& {$2\times 2$} & {N/A} \\
  \hline
  Quadrature rule& {$3\times3$} & {N/A} \\
   \hline
   Epochs & \multicolumn{2}{c|}{150,000} \\
   \hline
  \makecell{Training time: \\ {[s]}/1000 epochs}&   2.25  & 30.24\\
  \hline
    Error & \makecell {$e_0$: 1.44e-2\\ $e_1$: 1.37e-2}  & \makecell {$e_0$: 8.02e-2 \\ $e_1$: 2.84e-1 } \\
   \hline
\end{tabular}
\caption{Hyperparameters and training results of the V-NIM/h and the mixed-variable PINN for 2D elasticity problem, where the FEM reference solution is used for calculating the errors.}
\label{tab:setting_plate}
\end{table}

\begin{figure}[htb]
	\centering
	\includegraphics[angle=0,width=0.8\textwidth]{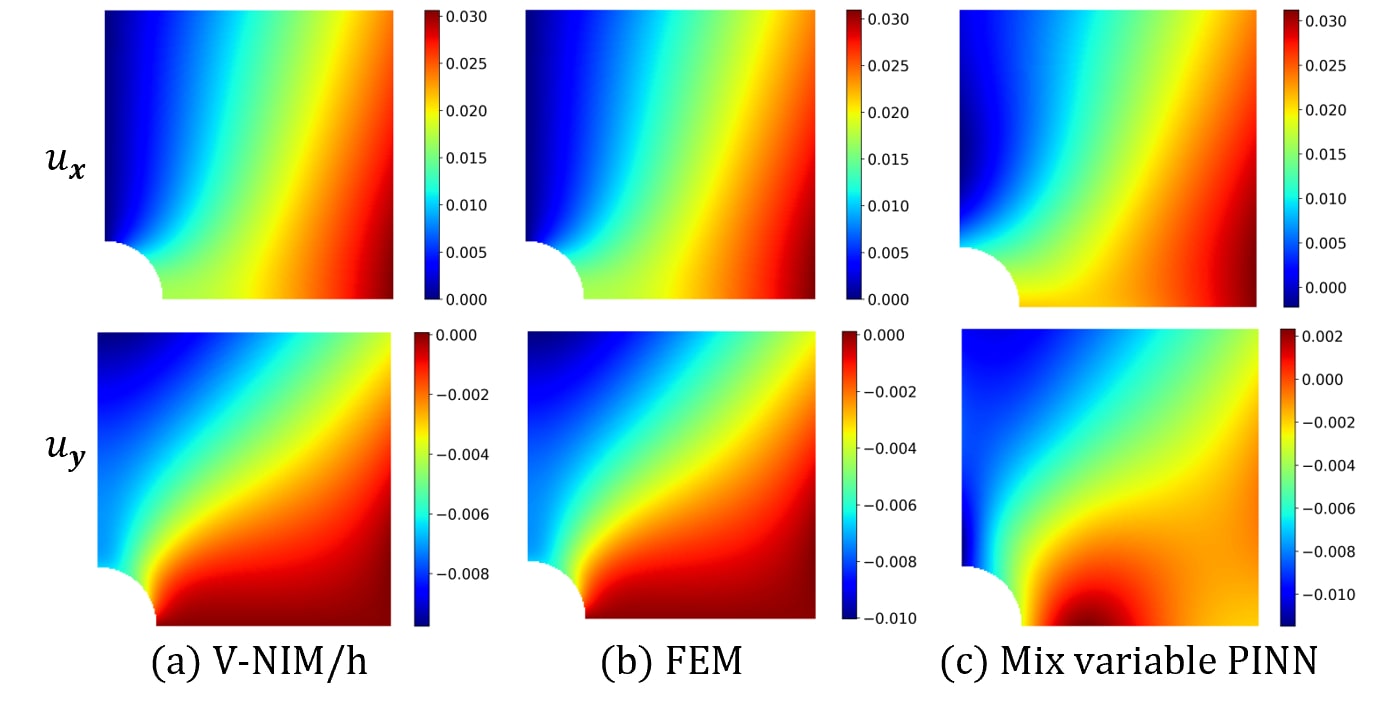}
	\caption{Comparison of the approximated displacement computed by (a) V-NIM/h, (b) FEM and (c) Mixed-variable PINN.}
    \label{fig:u_compare}
\end{figure}

\begin{figure}[htb]
	\centering
	\includegraphics[angle=0,width=0.8\textwidth]{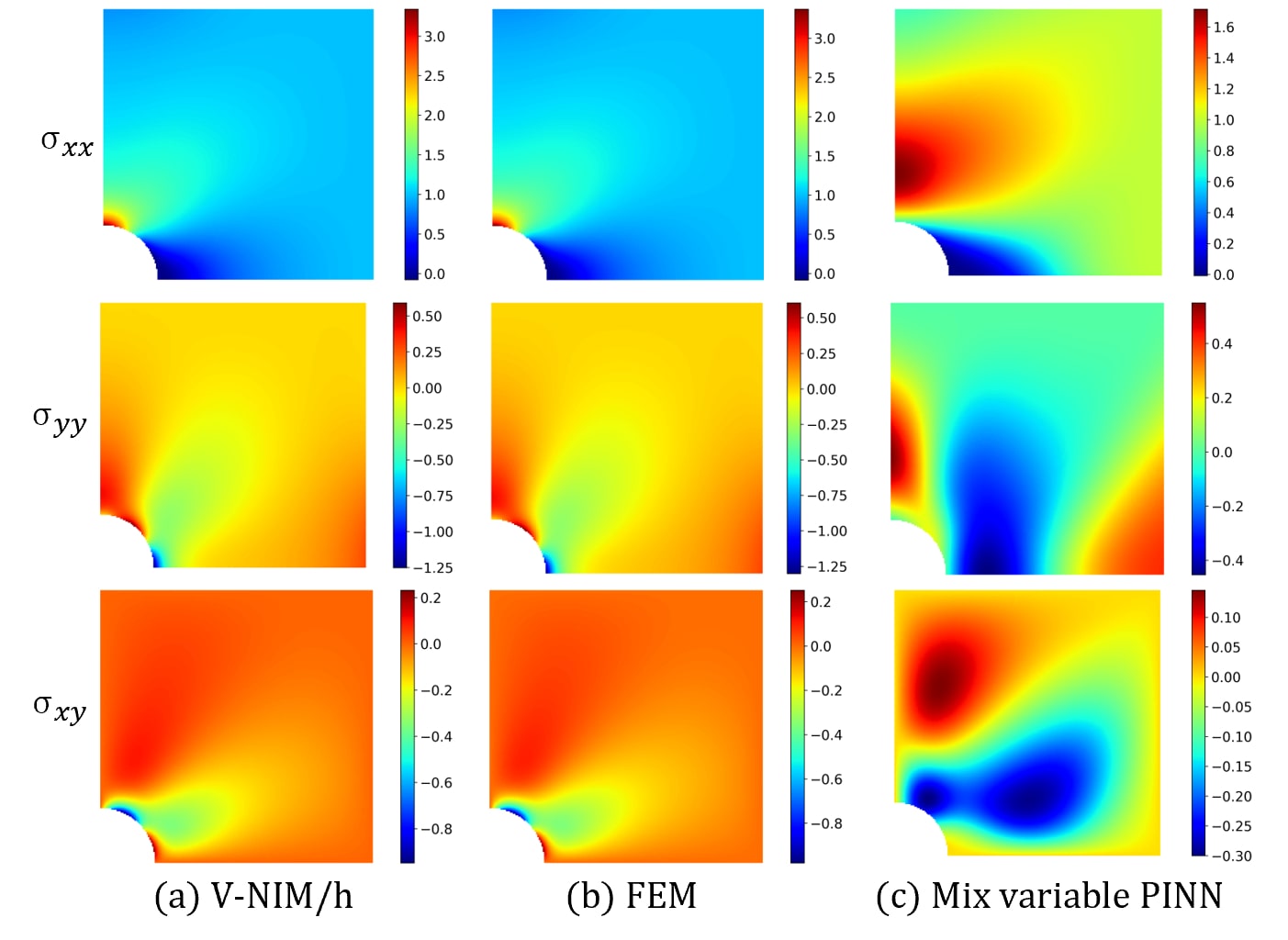}
	\caption{Comparison of the approximated stress components $\sigma_{xx}$, $\sigma_{yy}$ and $\sigma_{xy}$ computed by (a) V-NIM/h, (b) FEM and (c) Mixed-variable PINN.}
    \label{fig:s_compare}
\end{figure}
\begin{figure}[htb]
	\centering
	\includegraphics[angle=0,width=1.0\textwidth]{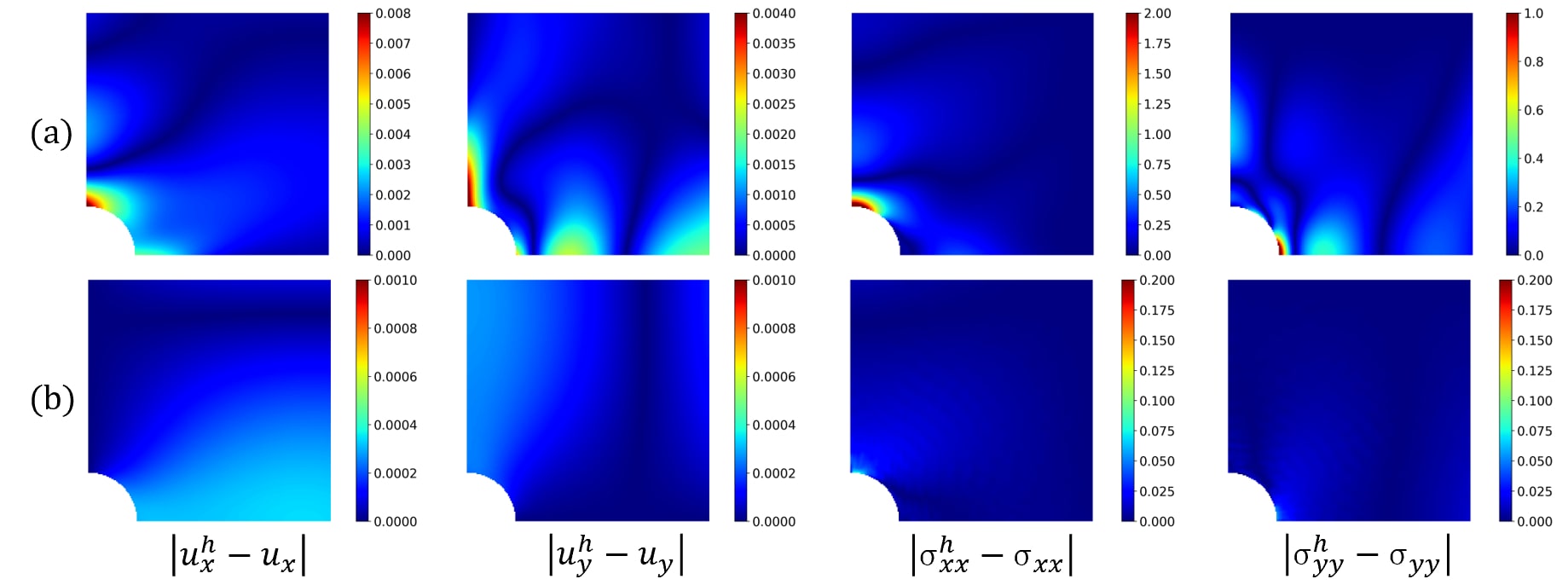}
	\caption{Comparison of the point-wise error of displacement and stress components obtained by (a) Mixed-variable PINN, (b) V-NIM/h.}
    \label{fig:err_cylinder}
\end{figure}

The V-NIM/h method exhibits remarkably high training efficiency, as shown in Table \ref{tab:setting_plate}, with a training speedup of nearly 15 times compared to the mixed-variable base PINN method (i.e., 2.25s vs. 30.24s for every 1000 epochs of training). 
Apart from the training time, Table \ref{tab:setting_plate} shows the comparison of $e_0$ and $e_1$ errors against the FEM reference solution, revealing approximately 5 times higher accuracy in V-NIM/h compared with PINN. 
This illustrates the effectiveness of the end-to-end differentiation capacity within  the proposed variational framework, as well as the enhanced efficiency and accuracy 
achieved through the use of NeuroPU for computing approximations and spatial gradients.

The approximated displacement and stress distributions generated by V-NIM/h, PINN and FEM methods are visualized in Figures \ref{fig:u_compare} and \ref{fig:s_compare}, respectively. 
It is observed that V-NIM/h consistently yields agreeable results with the reference solution obtained by the FEM method, while the PINN model is not capable of capturing the stress concentration around the notch region, leading to a less satisfactory solution. 
The point-wise error of the displacement ($u_x$ and $u_y$) and two stress fields ($\sigma_{xx}$ and $\sigma_{yy}$) are also portrayed in Figure \ref{fig:err_cylinder}. It shows that the errors of PINN are normally $5 \sim 10$ times larger than the corresponding errors produced by V-NIM/h.
The improvement of V-NIM is particularly pronounced in the stress field, which further confirms the proposed method offers enhanced approximation capability in the higher-order derivatives of solution.

\begin{figure}[htb]
	\centering\includegraphics[angle=0,width=1.0\textwidth]{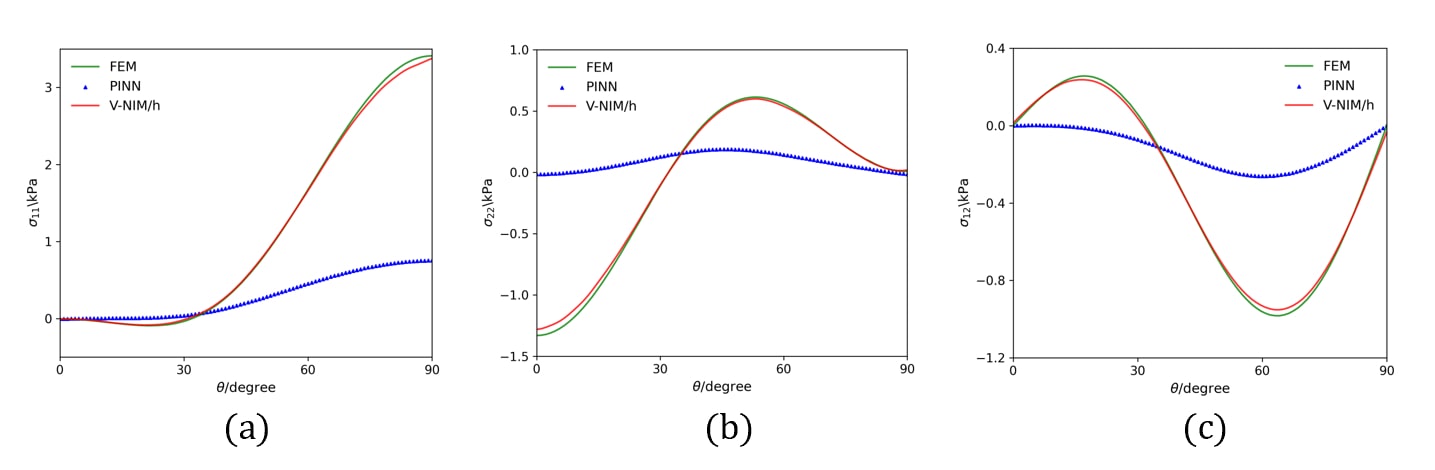}
	\caption{Comparison of the stress distribution (a) $\sigma_{xx}$, (b) $\sigma_{yy}$ and (c) $\sigma_{xy}$ on the notched surface obtained by the PINN, V-NIM/h and FEM methods.}
    \label{fig:notch}
\end{figure}

Taking a closer observation, the stress distributions around the notch are plotted in Figure \ref{fig:notch}, where the reference FEM, PINN, and V-NIM/h are compared. 
It is evident that the solutions obtained by V-NIM/h method closely align with the reference solution, demonstrating its advantage of accuracy in critical regions. 
In contrast, the results obtained through the PINN method show significant deviations from the reference solution, failing to capture the steep changes of stress on the notch along with the degree. This could be due to the insufficient approximation of the network model and the training difficulty associated with the elasticity problem.

\subsection{Surrogate modeling for parameterized elliptic PDE}\label{sec:surrogate}
In this example, we aim to investigate the applicability of the NIM method for surrogate modeling. 
Let us consider a parameterized elliptic PDE \cite{grepl2007efficient} defined in $\Omega = [0,1] \times [0,1]$
\begin{equation}
-\nabla^2 u(x, y)+\frac{\mu_1}{\mu_2}\left(e^{\mu_2 u}-1\right)=100 \sin (2 \pi x) \sin (2 \pi y)
\end{equation}
and the associated Dirichlet (essential) boundary condition defined at $\Gamma = \partial \Omega$ is
\begin{equation}
u(0, y)=u(1, y)=u(x, 0)=u(x, 1)=0
\end{equation}
where $\mu_1$ and $\mu_2$ are the system parameters, with $\left(\mu_1, \mu_2\right) \in \mathcal{D}=[0.01,10]^2$. 
Following the network architecture of NeuroPU shown in Figure \ref{fig:structure}, we input $(\mu_1, \mu_2)$ into the neural network. The resulting outputs are the respective $N_h$ nodal coefficient functions obtained after passing through hidden layers structured as $4 \times [40]$, representing the surrogate model associated with the two parameters.

We consider the V-NIM model with the cubic B-spline test function (V-NIM/c), where the computation domain is discretized using 441 uniform nodes, with an equal number of subdomains, i.e., $N_h=N_\mathcal{T}=441$. The penalty number for essential boundary enforcement is set as 10. The training of the V-NIM surrogate model is conducted by 7000 epochs, utilizing an Adam optimizer with a learning rate that decreases from 1e-3 to 1e-4.
The normalized support size and subdomain size are given by $\bar a=2.5$ and $\bar r=1.5$. 

To evaluate the extrapolation performance of the V-NIM surrogate, 121 uniformly distributed parameter points (i.e., $N_{\mu} =121$) within $\left(\mu_1, \mu_1\right) \in [0.01,6]^2 \subset \mathcal{D}$ are used for training the V-NIM model \eqref{eq:loss_v}, and then the trained model will be tested on the parameter set $\left(\mu_1, \mu_1\right)=[10,10]$ that is outside the range of training set.

\begin{figure}[htb]
	\centering
\includegraphics[angle=0,width=0.8\textwidth]{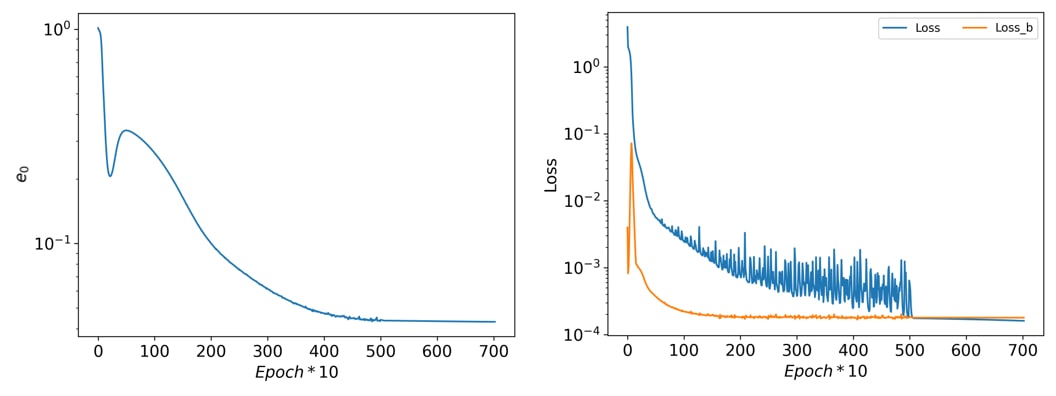}
\caption{The evolution of $L^2$ error (Left) and the $MAE$ loss function (Right) during training the V-NIM/c surrogate. The loss term associated with the Dirichlet boundary condition is also provided.}
    \label{fig:para_loss}
\end{figure}

\begin{figure}[htb]
	\centering
 \includegraphics[angle=0,width=1.0\textwidth]{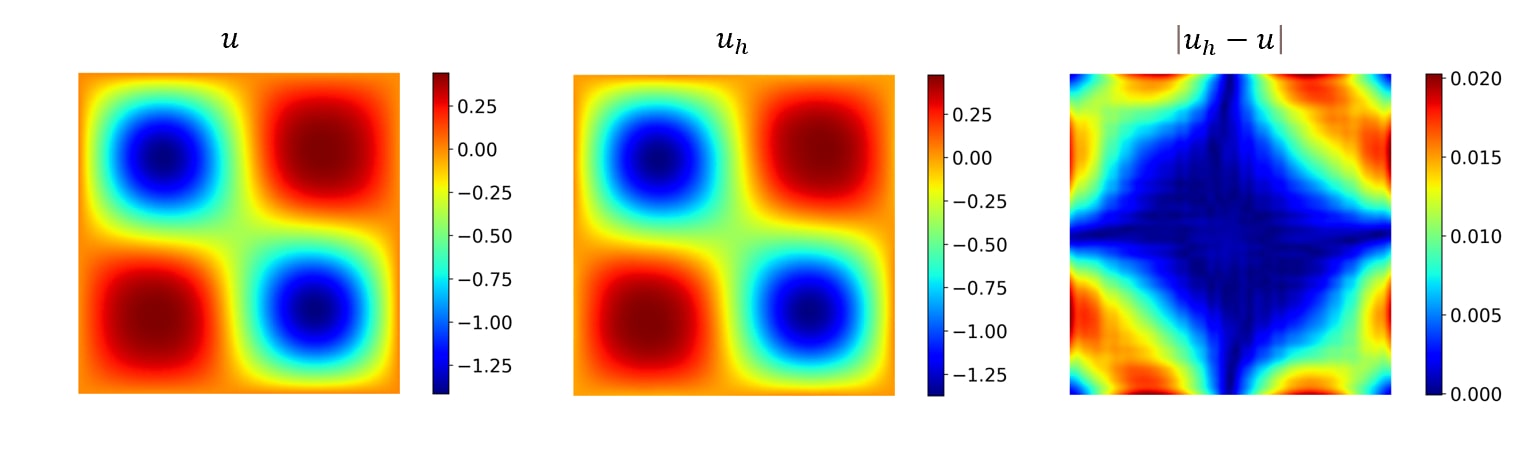}
	\caption{(Left) The reference solution, (Middle) predicted solution by the V-NIM/c surrogate; (Right) the point-wise error. The parameter testing point is $\left(\mu_1, \mu_2\right)=[10,10]$.}
    \label{fig:pde_err}
\end{figure}

Figure \ref{fig:para_loss} depicts the loss evolution during the training process of the V-NIM surrogate using the Adam optimizer for 7,000 epochs. It shows that the V-NIM model converges well after about 5,000 epochs and the boundary condition is satisfied adequately.
The prediction by V-NIM at the testing parameter point $\left(\mu_1, \mu_2\right)=[10,10]$ is given in Figure \ref{fig:pde_err}. It can be seen from the error distribution that the higher errors locate at boundary areas and the regions where large gradients appear. However, the maximum point-wise error of V-NIM is less than 0.02. 
Overall, the result exhibits a desirable agreement with the reference solution despite being outside the training set. This demonstrates the ability of the NIM method for accurate extrapolative prediction beyond the training region. 

\begin{figure}[htb]
	\centering
    \includegraphics[angle=0,width=0.6\textwidth]{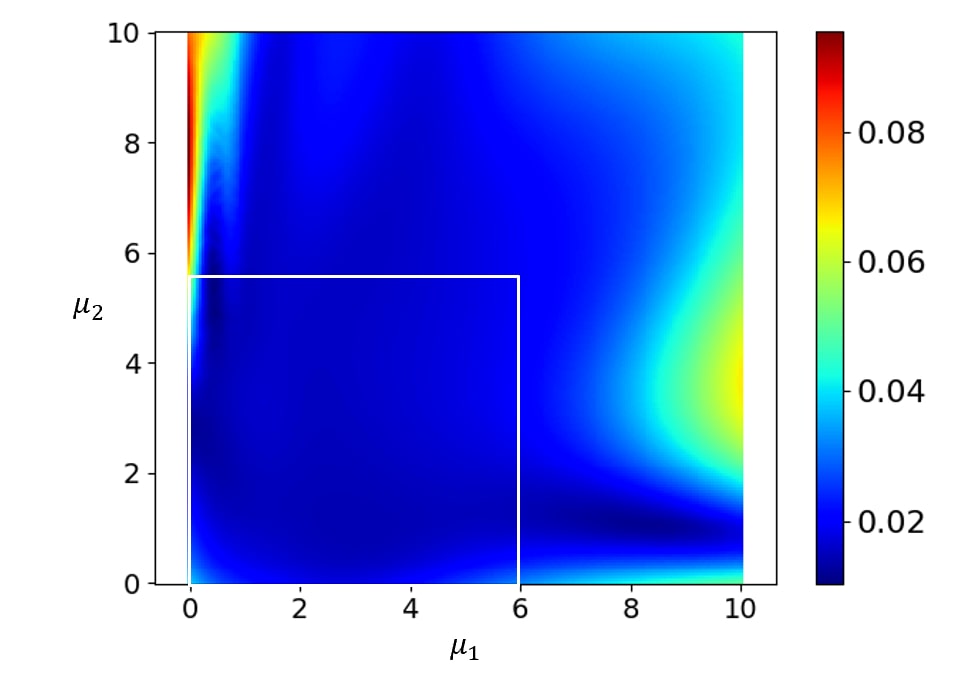}
	\caption{The distribution of $e_0$ error of the V-NIM surrogate over the parameter space $\left(\mu_1, \mu_1\right) \in \mathcal{D}$.}
    \label{fig:prediction}
\end{figure}

Additionally, Figure \ref{fig:prediction} presents the $e_0$ error of the predicted solution for $\left(\mu_1, \mu_2\right) \in \mathcal{D}$, where the training set is enclosed within a white box. The remaining area represents the extrapolation region of the surrogate model, illustrating the capability of V-NIM in surrogate modeling.
It is noteworthy that once the NIM surrogate model is trained, the predictions for other parameter points can be generated in real-time with minimal additional computational cost.  This surrogate modeling capacity of the NIM framework offers a unique advantage compared to conventional numerical solvers that require repeated full computation for each case.

\section{Numerical Results: Time-Dependent Problem}\label{sec:result_ade}
To showcase the applicability of the NIM method to dynamics problems, we consider a time-dependent advection-diffusion (AD) equation \cite{he2021physics,kharazmi2021hp} defined in computation domain $(x, t) \in \Omega \times \Omega_t$
\begin{equation}
\left\{\begin{array}{c}
u_{, t}+a u_{, x} =\kappa u_{, x x}, \quad \text{in} \quad \Omega \times\Omega_t \\
u(\pm 1, t) = 0 \\ 
u(x, 0)  =  u_0 (x) = -\sin (\pi x)
\end{array}\right.
\label{eq:ade}
\end{equation}
where $\Omega=[0, 1]$, $\Omega_t=[0, 1]$, the Dirichlet (Essential) boundary is $\Gamma_g = \{(x,t) | x = \pm 1, t \in \Omega_t\}$, the initial boundary is $\Gamma_0 = \{ (x,t) | x \in \Omega, t = 0 \}$, and 
$a=1$ and $\kappa=0.1/\pi$ represent the advection velocity and the diffusivity coefficient, respectively.  The analytical solution of Eq. \eqref{eq:ade} is given in \cite{mojtabi2015one} with infinite series summation.

\begin{table}[htb]
\centering
\small
\begin{tabular}{|c|c|c|}
  \hline
Methods & V-NIM/c & \textit{hp}-VPINN  \\
\hline
Trial functions & NeuroPU & Neural network \\
 \hline
 Test functions & Cubic B-spline  & Legendre polynomials ($P_n(x),n=1,...,5$)\\
 \hline
  $N_h$ &  41 & N/A  \\
 \hline
  Subdomain size $ r$ & $1.5h$ & $0.25$\\
 \hline
  Support size $a$ & $2.5h$ & N/A\\
  \hline
 $N_\mu$ & 41 & N/A \\
  \hline
 Subdomain $N_{\mathcal{T}}$& 41 (1D) & $4 \times 4$ (2D)\\
  \hline
  $\alpha_1, \alpha_2 $ & 1,1 & 10, 10 \\
  \hline
  Quadrature rule& 15*1 (1D) & 10*10 (2D) \\
  \hline
  Epochs & 100000 & 150000\\
  \hline
\end{tabular}
\caption{Hyperparameters of the V-NIM and \textit{hp}-VPINN for the advection-diffusion equation. The support size $a$ and the subdomain size $r$ are calculated by $a=\bar a h=1.5 /20=0.075$ and $r=\bar r h= 2.5/20=0.125$, where $h=1/20$ is the characteristic nodal distance.}
\label{tab:ADE_parameters}
\end{table}

By defining the temporal variable as a system parameter input, the NeuroPU approximation \eqref{eq:nmm} for this problem can be rewritten as
\begin{equation}
\hat{u}^h(x,t;\bm \theta) =  \sum_{I \in \mathcal{S}_x} \Psi_I({x}) \hat{ d}_I(t;\bm \theta)
\label{eq:nmm_ade}
\end{equation}
It is noted that, aided by NeuroPU approximation, the NIM method avoids the necessity of applying semi-discretization to the spatiotemporal domain that is commonly used in numerical schemes 
for time-dependent problems~\cite{leveque2007finite,hughes2012finite}. 
Instead, NIM is able to directly approximate the solution as a continuous function of temporal variable.
While it is not within the scope of this study, this continuous representation opens up a potential avenue to deal with time-dependent observation data for inverse dynamics problems. 
On the other hand, NIM simplifies the integration process by decoupling the spatial and temporal domains,
unlike the two-dimensional spatiotemporal integration required for the weak form-based PINN methods~\cite{kharazmi2021hp}.

For demonstration, we utilize the V-NIM method with cubic B-spline function as the test functions, i.e., V-NIM/c method, to solve this problem. With the employment of Eq. \eqref{eq:nmm_ade}, the local variational residual for the AD equation \eqref{eq:ade} over $\Omega_s$ can be derived as
\begin{equation}
\begin{aligned}
\mathcal{R}_s^h=&\int_{\Omega_s} v \left(\hat{u}_{, t}^h+a \hat u_{, x}^h-\kappa \hat u_{, x x}^h\right) d \Omega \\ = & \kappa \int_{\Omega_s} v_{, x} \hat u_{, x}^h d \Omega -\kappa \int_{\Gamma_{s u}} v \hat u_{, x}^h n_x d \Gamma +\int_{\Omega_s} v\left(\hat u_{, t}^h+a \hat u_{, x}^h\right) d \Omega 
\end{aligned}
\label{eq:residual_ade}
\end{equation}
where the divergence theorem is applied and the boundary integral term on $L_s$ is canceled due to the boundary vanishing property of the cubic B-spline function employed as test functions (refer to Section \ref{sec:testfunction}). The corresponding loss function modified based on \eqref{eq:loss_v} can be formulated as
\begin{equation}
\begin{aligned}
\mathcal{L}^V (\bm \theta) &=\frac{1}{N_{\mu}N_\mathcal{T}}\sum_{j=1}^{N_{\mu}}\sum_{s=1}^{N_\mathcal{T}}\left\|{\mathcal{R}}_s^{h}(t_j) \right\|^2 +\frac{\alpha_1}{N_{\mu}}\sum_{j=1}^{N_{\mu}}\left\|\hat{u}^h(\pm 1,t_j)-\bar u(\pm 1, t_j)\right\|^2  \\ 
&+ \frac{\alpha_2}{N_\mathcal{T}}\sum_{s=1}^{N_\mathcal{T}}\left\|\hat{u}^h(x_s,0)-{u_0}(x_s)\right\|^2
\end{aligned}
\label{eq:loss_ade}
\end{equation}
where $N_{\mu}$ is considered as the number of sampling points in the parameter space, namely, the temporal space $\Omega_t$. 

As shown in Table \ref{tab:ADE_parameters}, for the setting of V-NIM/c, $N_h = 41$ uniformly distributed nodes and $N_{\mathcal{T}} = 41$ subdomains are utilized for meshfree discretization in the physical domain $\Omega=[-1,1]$, whereas $N_{\mu} = 41$ sampling points along the temporal dimension $\Omega_t=[0,1]$ are used to train the neural network of NeuroPU approximation. The normalized size of subdomains and quadratic trial shape functions are set as $\bar r=1.5$ and $\bar a=2.5$, respectively.
For comparison, we also conduct simulations using the \textit{hp}-VPINN method \cite{kharazmi2021hp}, where the associated hyperparameters of \textit{hp}-VPINN are kept the same as Figure 16 in the reference \cite{kharazmi2021hp} and summarized in Table \ref{tab:ADE_parameters}. 
We would like to emphasize that achieving a perfectly fair comparison is nontrivial, if not impossible, due to the introduction of NeuroPU and the local variational form. These elements render the proposed V-NIM a fundamentally distinct approach when compared to \textit{hp}-VPINN. 
For instance, we have noticed that \textit{hp}-VPINN necessitates using a set of test functions upto 5th order polynomials as well as a large number of training epochs to attain sufficient accuracy, which however are not required in the NIM solver. 
Hence, the following comparative study should be regarded as an observational analysis, recognizing the inherent differences in the methodologies being compared.

The numerical tests with regard to different sizes of neural networks are conducted to show the accuracy and robustness of the proposed method V-NIM/c and \textit{hp}-VPINN.
As can be seen from the $MAE$ and $e_0$ errors in Table \ref{tab:ADE_results}, an enhancement in accuracy is attained as a larger neural network is adopted in the V-NIM method, with increasing training cost (see the "Time/100 epochs" column in Table \ref{tab:ADE_results}) as expected. 
This demonstrates a larger network can provide a better approximation capacity to capture the time-dependent behaviors. 
Due to the hybrid approximation property by integrating high-order meshfree shape functions and neural networks, the search space in V-NIM is greatly reduced compared to the conventional method (e.g., \textit{hp}-VPINN) purely approximated by neural networks.
Consequently, the proposed V-NIM is capable of attaining a preferable accuracy as \textit{hp}-VPINN but only with a much smaller network ($1 \times [10]$ vs. $3 \times [20]$), leading a fraction of training cost (as shown in Table \ref{tab:ADE_results}).
In addition, rather than relying on expensive AD to calculate high-order derivatives,
the introduction of NeuroPU approximation in NIM enables the pre-computed spatial gradients that singly operated on the shape functions~\eqref{eq:grad_nma}.
This also contributes to a reduction in the computational complexity. 

\begin{figure}[htb]
	\centering
	\includegraphics[angle=0,width=1.0\textwidth]{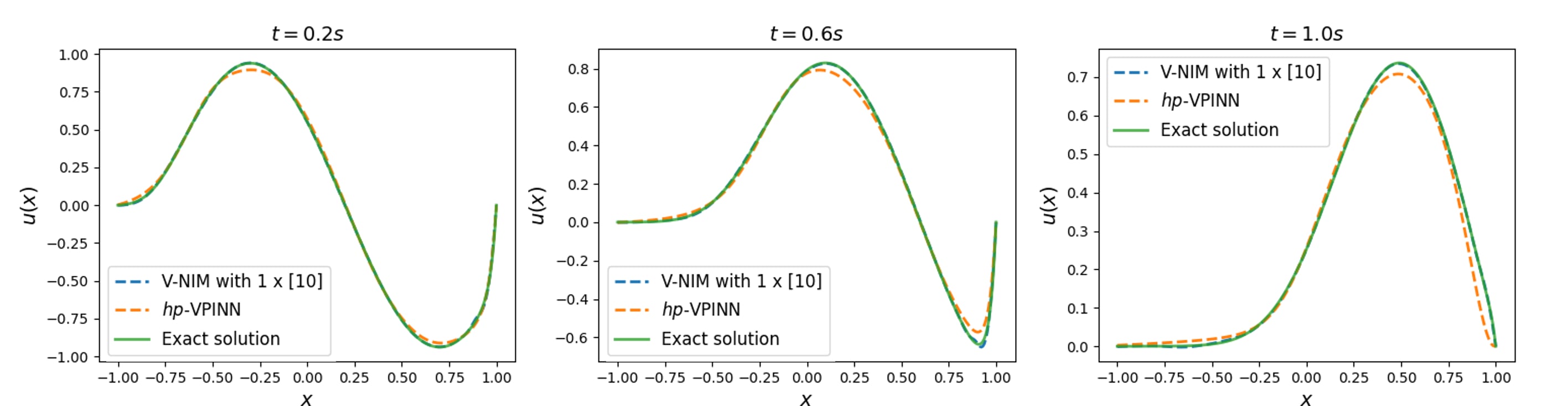}
	\caption{The snapshots of analytical solution, and approximation solutions obtained by V-NIM/c and \textit{hp}-VPINN methods at $t=0.2s, 0.6s$ and $1.0s$.}
    \label{fig:snap_ade}
\end{figure}
\begin{figure}[htb]
	\centering
	\includegraphics[angle=0,width=1.0\textwidth]{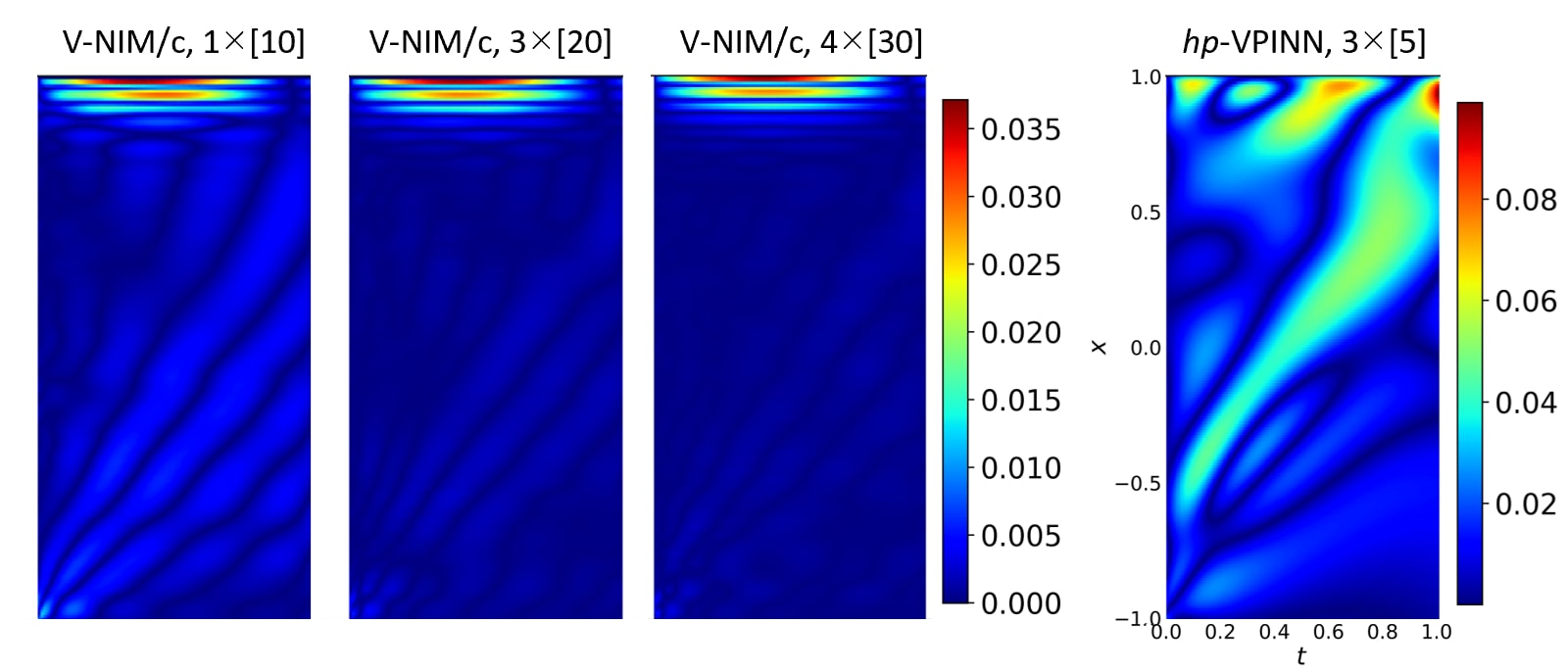}
    \caption{Comparison of the point-wise errors of the approximate solution obtained by V-NIM/c and \textit{hp}-VPINN. The horizontal and vertical axes denote the time and space coordinates, respectively.}
    \label{fig:ade_err}
\end{figure}

The snapshots of the solutions of the V-NIM/c and \textit{hp}-VPINN methods at $t=0.2s, 0.6s$ and $1.0s$ are plotted in Figure \ref{fig:snap_ade}, which evinces that the V-NIM/c solution agrees well with the analytical solution, while for the \textit{hp}-VPINN result a small deviation is observed in the region where the solution has relatively large magnitudes. This implies the enhanced approximation of V-NIM for the sharply changing solution and higher-order derivatives.
Figure \ref{fig:ade_err} shows the comparison of point-wise errors of \textit{hp}-VPINN and V-NIM/c over the whole spatial-temporal domain, with different sizes of neural networks. The maximum point-wise error in V-NIM/c solutions with different sizes of neural networks consistently remains under 0.04, with only a minor error observed at the top edge. In contrast, the error snapshot of the \textit{hp}-VPINN solution exhibits larger errors over most of the domain. It is also noted that 50,000 more epochs are used in training the \textit{hp}-VPINN model to obtain a satisfactory result.

\begin{figure}[htb]
	\centering
	\includegraphics[angle=0,width=1.0\textwidth]{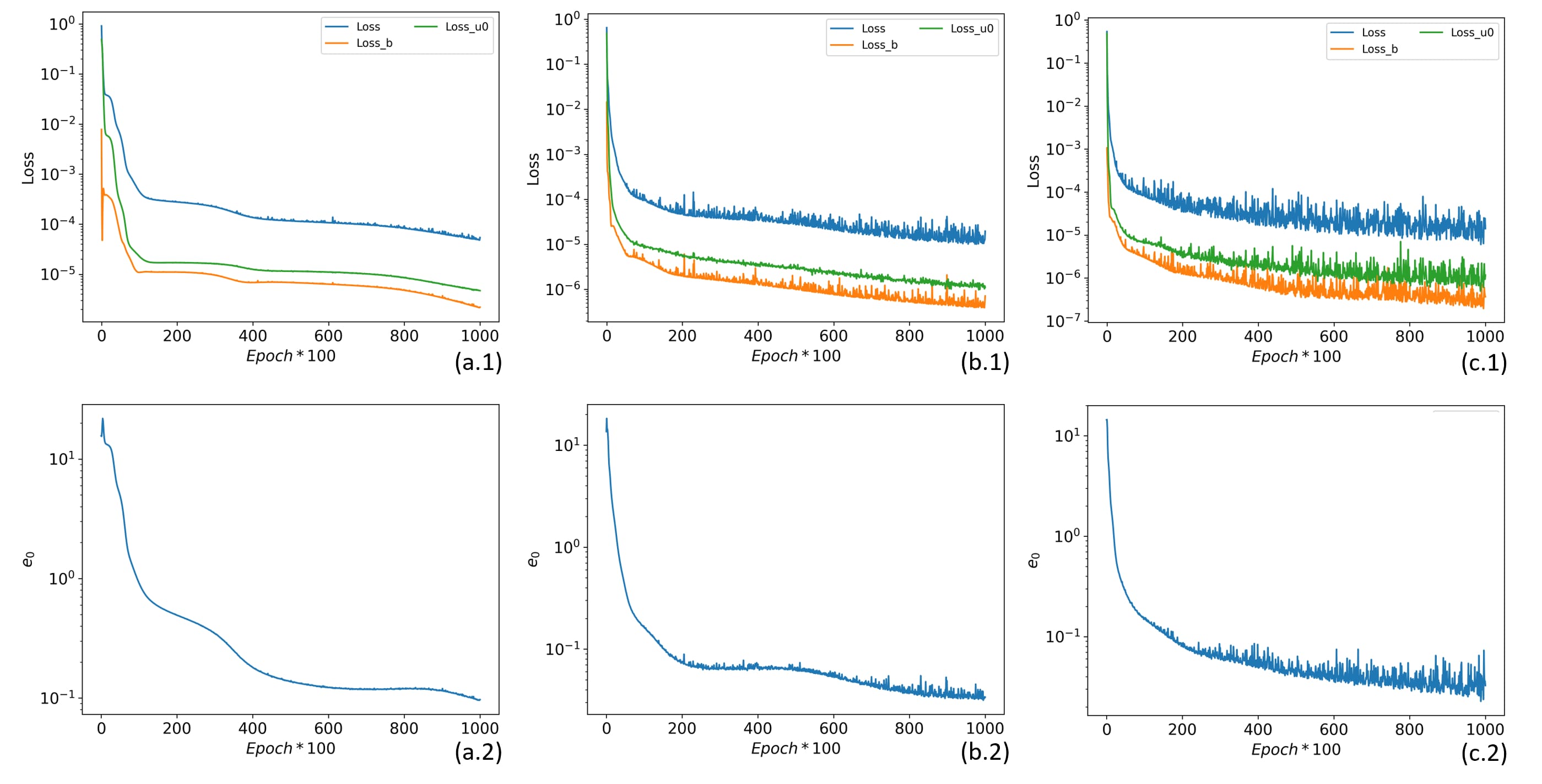}
	\caption{The evolution of training loss and $e_0$ error generated by V-NIM/c using different neural network architectures: (a) $1 \times [10]$, (b) $3 \times [20]$ and (c) $4 \times [30]$.}
    \label{fig:ade_loss}
\end{figure}

As supplementary, we provide the evolution of training loss of V-NIM/c using neural network with dimensions $1 \times [10]$, $3 \times [20]$ and $4 \times [30]$ in Figure \ref{fig:ade_loss}, which showcases the stable convergence property of the NIM method.

\begin{table}
\centering
\begin{tabular}{lllll}
\hline Methods & Neural networks &Time [s]/100 epochs & $MAE$  & $e_0$ \\
\hline
V-NIM/c & $1 \times [10]$ & 0.98 & 2.64e-03 &  9.59e-02 \\
V-NIM/c & $2 \times [10]$ & 1.16 & 2.04e-03 &  7.64e-02 \\
V-NIM/c & $3 \times [20]$ & 1.37 & 1.37e-03 &  3.24e-02 \\
V-NIM/c & $4 \times [30]$ & 1.67 & 1.17e-03 &  2.59e-02 \\
\textit{hp}-VPINN & $3 \times [5]$ & 1.23 & 1.81e-02 &  7.24e-01 \\
\textit{hp}-VPINN & $3 \times [20]$ & 1.91 & 1.69e-02 &  5.83e-01 \\
\hline
\end{tabular}
\caption{Comparison of V-NIM/c and \textit{hp}-VPINN under different sizes of neural networks. The model parameters can be referred to Table \ref{tab:ADE_parameters}.}
\label{tab:ADE_results}
\end{table}

\section{Conclusion}\label{sec:conclusion}
In this study, we present a novel framework, NIM, as a differentiable programming-based AI methodology to solve a variety of computational mechanics problems.
The integration of the numerically discretized system using meshfree methods with deep neural networks encoded in differentiable computation graphs enables the end-to-end training of the entire hybrid model to seek approximate solutions of the PDEs. 
The main characteristic of NIM is the hybrid approximation scheme, NeuroPU, which interpolates DNN representations with meshfree basis functions based on the PU concept, is introduced to enhance the approximation accuracy and computational efficiency. 
As an example, the reproducing kernel (RK) shape function is employed in the NeuroPU approximation as it permits arbitrary accuracy and smoothness defined a priori and it is well-suited for meshfree discretization. 
Thanks to the interpolation property of NeuroPU, the size of neural networks and subsequently the number of sampling points required to train the NIM solution can be significantly reduced while achieving satisfactory accuracy.

Within the proposed NIM framework, we propose two meshfree solvers, S-NIM and V-NIM.
While S-NIM presents a straightforward solution procedure by using strong-form PDEs in the loss function, 
our particular interest lies in V-NIM, which is built on the variationally consistent formulation, so that the required regularity of the approximate solution reduces.
To achieve this, we introduce a local Petrov-Galerkin approach that constructs the loss function of V-NIM using local residuals defined on overlapping subdomains.
This embedded meshfree property eliminates the need for costly conforming mesh generation and enables efficient batch training.
The variational nature of V-NIM offers the flexibility of using various test functions. Two types of test functions including Heaviside step function and cubic B-spline function, are employed for investigation in our study.

The merits and effectiveness of the NIM solvers have been demonstrated through various numerical examples across static and dynamic problems in comparison with the classical FEM, standard PINN and variational PINN methods.
The results of static problems (Section \ref{sec:pro_poi} and \ref{sec:pro_ela}) show that the NIM methods utilizing the NeuroPU approximation significantly enhance both efficiency and accuracy in comparison to the PINN method. For instance, in the case of Poisson's problem, S-NIM exhibits half the $MAE$ errors and requires only 1/5 of the training time when using a similar number of sampling points compared to PINN. 
Attributing to the enhanced stability and consistency enforced by the local variational formulation, even greater improvements are achieved by V-NIMs (V-NIM/c, etc.), leading to nearly an order of magnitude higher accuracy while being 10 times faster than PINN. V-NIMs also demonstrate favorable convergence rates under different orders of NeuroPU shape functions.
In the elasticity problem, we also highlight the superior performance of V-NIM in approximating the stress field, a higher-order derivative of the solution, compared to the strong-form methods, such as S-NIM and PINN.

The versatility of the proposed differentiable method beyond conventional simulation techniques is also demonstrated in this study. Specifically, we highlight the desirable extrapolative ability and the real-time prediction of the V-NIM model for surrogate modeling of a parameterized elliptic PDE.
It is also shown that, by considering the temporal variable as inputs to the NeuroPU approximation, the NIM method yields more stable results while maintaining superior training efficiency compared to the \textit{hp}-VPINN method for the time-dependent (advection-diffusion equation) problem. 

In conclusion, we describe the NIM method as a differentiable meshfree solver, enabling end-to-end gradient-based optimization procedures for seeking solutions. 
This innovative method holds great promise for the development of next-generation physics-based data-driven solvers that offer a remarkable balance between accuracy and computational efficiency.
Additionally, it provides a versatile framework for data assimilation and adaptive refinement due to its meshfree nature.
It is also worth noting that NIM opens a new avenue for creating more efficient deep learning models through seamlessly blending shape functions that represent the finite discretized domain with DNNs that represent the problem-parameter space.
In the future, we plan to explore the performance of the NIM method in operator learning, inverse modeling, and diverse applications related to nonlinear material modeling.

\section*{Acknowledgment}\label{Ack}
This research was partially supported by Q.Z. He's Startup Fund
and Data Science Initiative (DSI) Seed Grant
at the University of Minnesota. 
The authors also acknowledge the Minnesota Supercomputing Institute (MSI) for providing resources that contributed to the research results reported within this paper.

\appendix
\section{}\label{sec:app_A}

For 2D elasticity associated with the NIM framework (Section \ref{sec:NIM}), the approximated displacement $\hat {\bm u}^h(\bm x)$ is given as
\begin{equation}
\hat {\bm u}^h(\bm x)=\sum_{I \in \mathcal{S}_x} \bm N_I(\bm x) \hat {\bm d}_I
\end{equation}
also the stress $\hat {\bm \sigma}^h(\bm x)$ and traction $\hat {\bm t}^h(\bm x)$ tensors are given as
\begin{equation}
\hat {\bm \sigma}^h(\bm x)=\sum_{I \in \mathcal{S}_x} \bm D \bm B_I(\bm x) \hat {\bm d}_I
\end{equation}
\begin{equation}
\hat {\bm t}^h(\bm x)=\sum_{I \in \mathcal{S}_x} \bm n \bm D \bm B_I(\bm x)\hat {\bm d}_I
\end{equation}
where $\bm N_I$ $\bm{B}_I$ $\bm n$, and $\bm D$ are given as

\begin{equation}
\boldsymbol{N}_I=\left[\begin{array}{cc}\Phi_{I, x} & 0 \\ 0 & \Phi_{I, y} \end{array}\right], \boldsymbol{B}_I=\left[\begin{array}{cc}\Phi_{I, x} & 0 \\ 0 & \Phi_{I, y} \\ \Phi_{I, y} & \Phi_{I, x}\end{array}\right], \quad \boldsymbol{n}=\left[\begin{array}{ll}n_x & 0 \\ 0 & n_y \\ n_y & n_x\end{array}\right]^T
\end{equation}
and
\begin{equation}
\boldsymbol{D}=\frac{\bar{E}}{1-\bar{\nu}^2}\left[\begin{array}{lll}1 & \bar{\nu} & \\ \bar{\nu} & 1 & \\ & & (1-\bar{\nu}) / 2\end{array}\right], \begin{cases}\bar{E}=E, \bar{\nu}=\nu & \text { (plane stress) } \\ \bar{E}=\frac{E}{1-\nu^2}, \bar{\nu}=\frac{\nu}{1-\nu} & \text { (plane strain) }\end{cases}
\end{equation}
with $E$ and $\nu$ being the Young’s modulus and Poisson’s ratio, 
respectively.
\newpage
\bibliographystyle{cas-model2-names} 

\bibliography{ref_diff.bib,ref_Meshfree.bib,ref_new.bib}

\end{document}